%% file: neurips_2026.tex
\documentclass{article}
\PassOptionsToPackage{numbers, sort&compress}{natbib}

 \usepackage[main,final]{neurips_2026}
 \usepackage{float}
 \usepackage{rotating}
 \usepackage{multirow}

\usepackage[utf8]{inputenc} 
\usepackage[T1]{fontenc}    
\usepackage{hyperref}       
\usepackage{url}            
\usepackage{booktabs}       
\usepackage{amsfonts}       
\usepackage{nicefrac}       
\usepackage{microtype}      
\usepackage{xcolor}         
\usepackage{xspace}
\usepackage{amsmath}
\usepackage{graphicx}
\usepackage{wrapfig}
\usepackage[normalem]{ulem}

\title{\ours: Rethinking Mass Spectrum Prediction as an Object Detection Problem}

\newcommand{\ours}{GLACIER\xspace}

\author{%
  Rui-Xi Wang $\quad$ Runzhong Wang\thanks{Corresponding to Runzhong Wang and Connor W. Coley.} \ $\quad$ Connor W. Coley\textsuperscript{*} \\
  Massachusetts Institute of Technology \\
  Cambridge, MA 02139 \\
  \texttt{\{rxwangtw, runzhong, ccoley\}@mit.edu}\\
}

\begin{document}
\maketitle 
\begin{abstract}
Predicting tandem mass spectra (MS/MS) from molecular structures represents a central task in analytical chemistry with direct relevance to clinical metabolomics, systems biology, and adjacent disciplines. 
In this work, we revisit the problem through the lens of object detection on molecular graphs. Molecular fragmentation, a central step in MS/MS prediction, can be approximated as detecting a set of subgraphs (i.e., fragments) and their associated spectral contributions. Existing fragment-based models follow a two-stage paradigm---first generating candidate fragments and then scoring them---analogous to two-stage R-CNNs in computer vision. Towards higher accuracy and faster inference, we introduce \ours, a single-stage transformer-based fragment detection neural network for molecular graphs. This unified formulation eliminates the need for candidate enumeration, enabling scalable and globally consistent modeling of molecular fragmentation. 
\ours is faster and more accurate than existing state-of-the-art by a significant margin, achieving 70.0\% and 69.7\% Top-1 retrieval accuracy with and without contrastive finetuning on the MassSpecGym dataset (from the previous SOTA of 64.0\%) and 52.5\% and 38.5\% respectively on the NIST'20 dataset (from 33.2\%). Furthermore, \ours provides nearly 8-fold inference speedup over our prior two-stage model. Code is available at \url{https://github.com/coleygroup/ms-pred}.
\end{abstract}

\section{Introduction}
Tandem mass spectra (MS/MS) are measured at scale in chemistry and biology campaigns to help elucidate the structures of unknown molecules. MS/MS prediction aims to model the fragmentation of molecules, where a given molecular graph yields a set of fragment ions with associated intensities. An accurate MS/MS predictor can generate pseudo-standard spectra for candidate molecules, accelerating chemical and biomedical discoveries as demonstrated in case studies including \citet{qiang2026language} and \citet{Wang2025.05.28.656653}. Existing approaches typically formulate this task as formula or fragment enumeration, relying on domain-specific heuristics~\citep{allen2015cfm-id,MAGMA} or multi-stage model architectures~\citep{goldman2023scarf,goldman2024iceberg,murphy2023efficiently,young2024fragnnet}. While effective, these formulations obscure a vital design perspective: MS/MS prediction can be viewed as an object detection problem by drawing analogy between visual objects and molecular graphs.

Specifically, MS/MS prediction can be formulated as taking a molecular graph as input, proposing a set of subgraphs (i.e., fragments) and predicting their properties (i.e., intensities). Since we can calculate the mass of each fragment and we have the predicted intensities, such a framework predicts mass spectrum as a set of $(\texttt{mass}, \texttt{intensity})$ pairs. This framework closely parallels object detection in computer vision---subgraph detection is the graph-based version of bounding-box detection, and the intensity regressor is analogous to the classifier head in computer vision. We also want to note that chemical rearrangement may occur in reality, and that a set of subgraphs does not fully reflect the physics of collision-induced dissociation~\citep{van2024spectroscopic}, but it can be viewed as a reasonable approximation, empirically, in the context of spectral prediction.


Such a perspective suggests that the methodological evolution of object detection offers a useful lens for rethinking MS/MS prediction. Early object detection approaches were dominated by two-stage pipelines, such as R-CNNs~\citep{girshick2014rcnn,girshick2015fastrcnn,ren2016fasterrcnnrealtimeobject,he2017maskrcnn}, which first predict region proposals and then classify and refine them by a separate neural network. Analogously, current state-of-the-art MS/MS predictors~\citep{wang2025neuralgraphmatchingimproves,Wang2025.05.28.656653,young2024fragnnet,goldman2023scarf} first propose candidate fragments using fragmentation simulators, and then use a separate model to score these candidates by predicting intensities. While effective, such approaches inherit key limitations of two-stage designs, including low inference throughput, complex training pipelines, and, specifically for MS/MS prediction, heuristic-based fragmentation.

In computer vision, these limitations motivated a shift towards one-stage architectures. One-stage detectors improved efficiency by predicting both bounding boxes and classification labels in a single pass, with further benefits brought by stabilized training~\citep{DBLP:journals/corr/RedmonDGF15,liu2016ssd}. More recently, transformer-based models such as DETR~\citep{carion2020endtoendobjectdetectiontransformers} reformulated detection as a set prediction problem, eliminating the need for heuristic components like anchor boxes and non-maximum suppression. These models enable unified one-stage training while capturing global dependencies through attention mechanisms.

\begin{wrapfigure}[17]{r}{0.45\columnwidth}
    \centering
    \vspace{-15pt}
    \includegraphics[width=\linewidth]{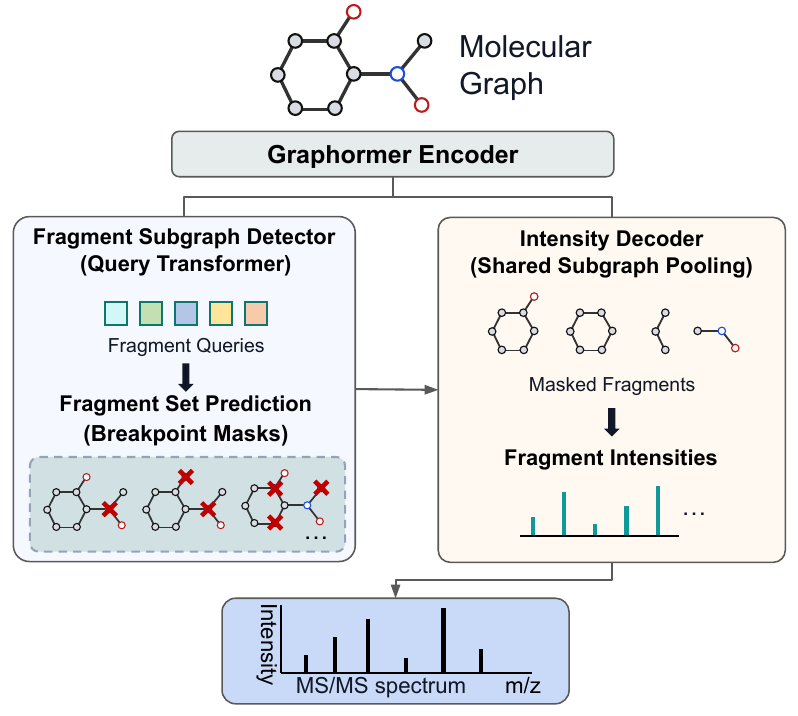}\vspace{-8pt}
    \caption{Overview of \ours as a one-stage MS/MS predicton neural network.}
    \label{fig:overview}
\end{wrapfigure}

Inspired by this progression, we envision that MS/MS prediction should undergo a similar paradigm shift. Instead of decomposing the problem into separate stages of candidate generation and scoring, we propose to directly model molecular fragmentation as a set prediction problem. As shown in Figure~\ref{fig:overview}, we present \underline{G}raph \underline{L}earning of \underline{A}tomic \underline{C}omponent \underline{I}nstances via \underline{E}nd-to-end \underline{R}ecognition (\ours), a transformer-based architecture that jointly proposes plausible subgraphs and predicts their intensities in the mass spectrum without relying on hand-crafted intermediate steps. 

Developing such an one-stage graph detection model also presents unique challenges, including predicting multiple breakpoints to generate complex fragments, enabling efficient subgraph feature pooling to maximize parameter efficiency, and mitigating the limitation of heuristic-based fragmentation. To address these challenges, we propose a principled one-stage formulation of MS/MS prediction that unifies fragment generation and intensity prediction with a transformer. This design generates fragments with various depths in a single forward pass, improves computational efficiency, and enables globally consistent modeling of molecular fragmentation. 

\textbf{Our contributions are summarized as follows:}

\textbf{1)} We incorporate a differentiable breakpoint predictor into \ours that supports multiple breakings. In contrast, existing single-stage MS/MS predictors only allow single-atom/bond fragmentation~\citep{nowatzky2025fiora}, resulting in limited coverage. Our approach matches the flexibility of two-stage models such as ICEBERG~\citep{Wang2025.05.28.656653} while retaining a one-stage formulation. This is achieved with multi-head predictors coupled with a differentiable constraint projection layer~\citep{WangICML23} that enables control over the number of broken atoms, and a separate branch to predict the cardinality. Inspired by DETR~\citep{carion2020endtoendobjectdetectiontransformers}, we utilize a query-based transformer to handle parallel prediction of multiple fragmentations; in training, the predicted breaking sites are mapped to the closest labeled patterns using the Hungarian algorithm.

\textbf{2)} In \ours, we utilize a shared Graphormer backbone~\citep{ying2021transformersreallyperformbad} and introduce an efficient subgraph pooling strategy that aggregates features of selected atoms directly from the global representation. This approach, referred to as dynamic embedding in prior work~\citep{wang2021combinatorial}, avoids redundant message passing over individual fragments, leading to improved inference speed and parameter efficiency. More broadly, while single-stage MS/MS prediction can benefit from a shared feature backbone, extracting subgraph-level features remains non-trivial. Naively applying message passing to each fragment in a one-stage model would undermine its theoretical efficiency, motivating our design.

\textbf{3)} In \ours, we propose a training strategy that uses fragment supervision with heuristic labels while progressively transitioning to an end-to-end objective. Specifically, we initialize the model with fragmentation labels generated by the MAGMa heuristic~\citep{MAGMA} and apply teacher forcing to learn spectral intensities, then gradually replace this signal with an end-to-end intensity objective. This design is motivated by a key challenge in MS/MS prediction: unlike visual object detection ground-truth fragment annotations are virtually always unavailable~\citep{van2024spectroscopic}. As a result, naive end-to-end training from scratch can be unstable. Our approach balances training stability with the goal of learning an accurate intensity predictor that remains robust to imperfect fragment assignments.

\textbf{4)} Extensive experiments on standard benchmarks demonstrate \ours's significant improvement in speed and accuracy. On the NIST'20~\cite{nist_database} dataset, compared to existing state-of-the-art, \ours improves top-1 retrieval accuracy from 33.5\% to 52.5\% on a random split and from 31.6\% to 46.0\% on a scaffold split. On the MassSpecGym dataset~\citep{bushuiev2024massspecgym}, \ours improves top-1 retrieval accuracy from 64.0\% to 70.0\% (mass challenge) and from 44.4\% to 49.9\% (formula challenge). The single-stage design also benefits efficiency, achieving a $\approx$8-fold speedup over a two-stage baseline~\citep{Wang2025.05.28.656653}.

\section{Related Work}
\textbf{Tandem Mass Spectra Prediction}.
Machine learning-based MS/MS prediction has enabled \textit{in silico} spectra generation directly from molecular structure. Several methods rely on kernel models and probabilistic fragmentation trees~\citep{allen2015cfm-id, wang2021cfm-id4, ruttkies2016metfrag}, while others leverage deep learning to capture complex fragmentation patterns~\citep{wei2019neims, young2024massformer}. Current best-performing approaches are dominated by two-stage, fragment-centric methods, such as ICEBERG~\citep{goldman2024iceberg, Wang2025.05.28.656653}, FraGNNet~\citep{young2024fragnnet}, and MARASON~\citep{wang2025neuralgraphmatchingimproves}, as well as a flow-based model, MolSpecFlow~\citep{Wang2026.01.28.702438}, aiming to better capture the generative nature of spectra. However, key limitations remain: the dominant two-stage pipeline introduces substantial computational overhead. 
While FIORA~\citep{nowatzky2025fiora} can perform one-step prediction with explicit fragmentation, it only supports single-step breakage and hence results in limited spectral peak coverage. These challenges motivate us to develop more efficient, single-stage models that better integrate fragmentation structure with downstream objectives.

\textbf{Object Detection in Computer Vision}.
Object detection aims at predicting bounding boxes and classes for visual objects. Early object detection models were two-stage, such as R-CNNs~\citep{girshick2014rcnn,girshick2015fastrcnn,ren2016fasterrcnnrealtimeobject,he2017maskrcnn}, which rely on region proposal networks followed by per-region classification and refinement, introducing architectural complexity and hand-designed components. One-stage detectors like YOLO~\citep{DBLP:journals/corr/RedmonDGF15} and SSD~\citep{liu2016ssd} improve efficiency, but still depend on heuristics such as anchor boxes and post-processing. More recently, DETR~\citep{carion2020endtoendobjectdetectiontransformers} reformulates detection as a set prediction problem using transformers, eliminating these components via bipartite matching and object queries. DETR-style models have been primarily developed for images and point clouds~\citep{misra2021endtoendtransformermodel3d}, while their extension to discrete graph-structured data remains underexplored. Unlike images, graphs lack a canonical spatial structure and exhibit irregular connectivity, making object definition ambiguous. In this work, we extend the DETR paradigm to molecular graphs by treating subgraph partitions as detection targets, enabling single-stage prediction of overlapping molecular fragments without predefined rules.

\section{The \ours Model}
\label{section:methods}
\begin{figure}[tbh!]
\includegraphics[width=\linewidth]{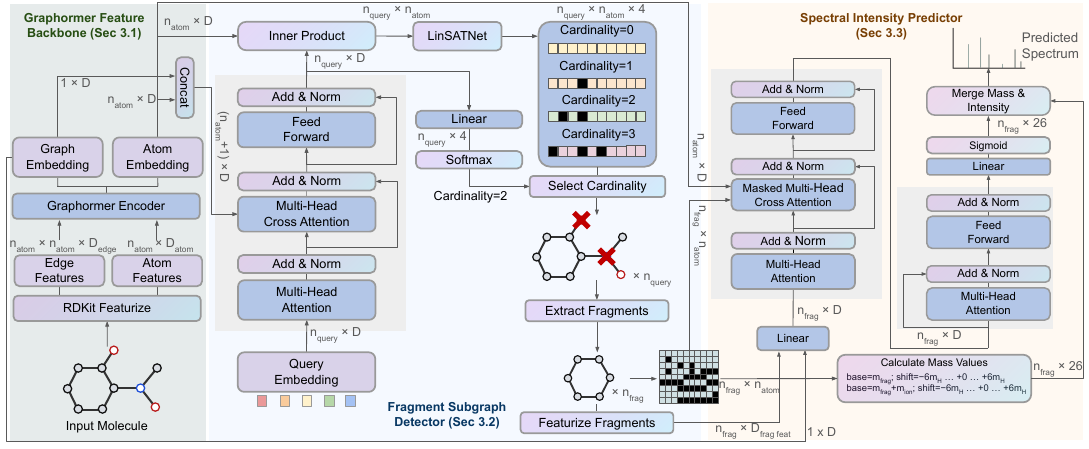}
    \vspace{-18pt}
    \caption{\ours architecture. Given a molecule as input, we first encode its structure using Graphormer to obtain contextualized graph and atom embeddings (Section~\ref{sec:backbone}). These embeddings are then used by the fragment subgraph detector to predict breakpoints and generate candidate fragment substructures (Section~\ref{section:breakpoint}). Finally, the spectral intensity predictor extracts subgraphs using Graphormer embeddings and predicts peak intensities for each predicted fragment (Section~\ref{sec:intensity}).}    
    \vspace{-5pt}
    \label{fig:overview-detailed}
\end{figure}

This section discusses three building blocks of \ours, as shown in Figure~\ref{fig:overview-detailed}, and the training loss.

\subsection{Graphormer Feature Backbone}
\label{sec:backbone}
We adapt Graphormer~\citep{ying2021transformersreallyperformbad} to encode molecules. Given a molecular graph, we represent each atom and bond as a feature fingerprint vector (see Appendix~\ref{section:features} for details). We then project these features, along with instrumental parameters such as adduct type (one-hot encoded) and collision energy (positional encoded), into node embeddings and edge-level attention biases via linear layers.

We process the resulting representations through multiple Graphormer layers together with a learned graph-level token. This produces node-level embeddings $\mathbf{N}\in \mathbb{R}^{n_\mathrm{atom}\times D}$ for $n_\mathrm{atom}$ nodes and a graph-level embedding $\mathbf{G}\in \mathbb{R}^{1\times D}$, where $D$ is the feature dimension. We subsequently use these embeddings with downstream modules for fragment generation and intensity prediction.

\subsection{Fragment Subgraph Detector}
\label{section:breakpoint}
\textbf{Graph Detection}.
The largest technical challenge of \ours is how to design a single-pass fragment detector with minimal additional enumerations to avoid significant inference burden on CPU. 
We formulate fragment prediction as an object detection problem on graphs, where each ``object'' corresponds to a fragment subgraph of the molecular graph. 
To achieve this, we adapt the transformer-based object detection framework, DETR~\cite{carion2020endtoendobjectdetectiontransformers}, to the graph domain. Following \citet{carion2020endtoendobjectdetectiontransformers}, we initialize a set of learnable query tokens $\mathbf{T} \in \mathbb{R}^{n_\mathrm{query} \times D}$. ``Object queries'' were proposed in DETR to provide a fixed number of anchors for predicted objects. In \ours, queries are used to predict atom-breaking patterns. $n_\mathrm{query}$ is the number of queries. $\mathbf{T}$ is initialized using the orthogonal initialization scheme~\cite{saxe2014exactsolutionsnonlineardynamics}. 
These query tokens cross-attend the concatenated node embeddings $\mathbf{N}$ and the graph-level embedding $\mathbf{G}$, and $[\cdot \;\|\; \cdot]$ means tensor concatenation along the feature dimension:
\begin{align}
    \bar{\mathbf{T}} = \mathrm{TransformerDecoder}(\mathbf{T}, [\mathbf{G} \;\|\; \mathbf{N}]).
\end{align}
$\bar{\mathbf{T}}$ is then used to compute a query-node affinity matrix $\mathbf{S} \in \mathbb{R}^{n_\mathrm{query} \times n_\mathrm{atom}}$, where $n_\mathrm{atom}$ is the number of heavy atoms and ${S}_{ij} = \bar{\mathbf{T}}_i \cdot \mathbf{N}_j$
measures the relevance between query $i$ and node $j$.

\textbf{Breakpoint Prediction}.
We convert the affinity matrix $\mathbf{S}$ into binary masks representing fragment subgraphs. 
However, directly predicting node-level masks is challenging in the mass spectrum setting, because even a single-atom difference in the mask will result in completely different mass values, and existing mass spectrum similarity metrics~\citep{li2021entropy} will treat that as unmatched. 
Instead, \ours predicts fragments by predicting the boundaries of the subgraphs, i.e., the breaking sites that induce the subgraphs.

Given an adjacency matrix $\mathbf{A} \in \{0,1\}^{n_\mathrm{atom} \times n_\mathrm{atom}}$ and a binary subgraph mask $\mathbf{m} \in \{0,1\}^{n_\mathrm{atom}}$, we define the ground-truth boundary (breakpoint) mask $\mathbf{b}^{gt} \in \{0,1\}^{n_\mathrm{atom}}$ as:
\begin{align}
    \label{equation:breakpoint}
    \mathbf{b}^{gt} = (\mathbf{A} \mathbf{m}) \odot (\mathbf{1} - \mathbf{m}),
\end{align}
where $\odot$ denotes element-wise multiplication. 
This formulation identifies nodes that are adjacent to the subgraph but not contained within it, analogous to the bounding box of an object in images.

With the observation that fragmentation events involve only a small number of bond breaks and to constrain the model dimension, we assume that the number of breakpoints is bounded by a hyperparameter $k$ (set to $k=3$ in our experiments). 
Given fragment subgraphs generated from the MAGMa heuristic~\citep{MAGMA}, we use Eq.~(\ref{equation:breakpoint}) to derive corresponding breakpoint patterns for supervision.

To predict breakpoint patterns, we project each affinity vector $\mathbf{S}_i \in \mathbb{R}^{n_\mathrm{atom}}$ into a set of cardinality-constrained vectors using the LinSATNet layer~\citep{WangICML23}:
\begin{align}
    \mathbf{C}_i = \big[\mathrm{LinSATNet}(\mathbf{S}_i, 0, \tau), \mathrm{LinSATNet}(\mathbf{S}_i, 1, \tau), \dots, \mathrm{LinSATNet}(\mathbf{S}_i, k, \tau) \big] \in \mathbb{R}^{(k+1)\times n_\mathrm{atom}},\notag
\end{align}
where $\mathbf{C}_{i,k^\prime}=\mathrm{LinSATNet}(\mathbf{S}_i, k^\prime, \tau)$ is a differentiable top-$k^\prime$ projection layer that enforces the cardinality by optimal transport, so that $\mathbf{C}_{i,k^\prime}$ satisfies $\mathbf{1} \cdot \mathbf{C}_{i,k^\prime} = k^\prime$, and $\mathbf{C}_{i,k^\prime} \in [0,1]^{n_\mathrm{atom}}$. Its closeness to binary $\{0,1\}$ is controlled by the temperature $\tau$. Therefore, each row of $\mathbf{C}_i$ is a soft top-$k^\prime$ activation of $\mathbf{S}_i$. The first row denotes the entire molecule and is always all 0s, corresponding to a cardinality of zero.

At the same time, for each $i$, we predict which cardinality to use, i.e., which row of $\mathbf{C}_i$ to select:
\begin{align}
    \mathbf{\Gamma} = \mathrm{Softmax}(\mathrm{MLP}(\bar{\mathbf{T}})) \in \mathbb{R}^{n_\mathrm{query} \times (k+1)},
\end{align}
where each entry in $\mathbf{\Gamma}$ represents the probability that a particular query corresponds to a breakpoint pattern with a particular cardinality between $0$ and $k$.

\textbf{Set Prediction Loss for Breakpoint Detection}.
Following the DETR framework~\citep{carion2020endtoendobjectdetectiontransformers}, our model produces a fixed-size set of $n_\mathrm{query}$ predictions, where $n_\mathrm{query}$ is chosen to be larger than the typical number of breakpoint patterns in a molecular graph. 
In training, it is treated as a permutation-invariant set prediction problem via bipartite matching.

We define the pairwise loss $\hat{\mathcal{L}}_{\text{breakpoint}}^{ij}$ between query $i$ and ground-truth breakpoint pattern $j$ as:
\begin{align}
\hat{\mathcal{L}}_{\text{breakpoint}}^{ij} = 
\mathrm{CrossEntropy}\!\left(
\frac{\mathbf{C}_{i,k_j^{gt}}}{\sum \mathbf{C}_{i,k_j^{gt}}},\;
\frac{\mathbf{b}_j^{gt}}{k_j^{gt}}
\right)
+ 
\mathrm{CrossEntropy}\!\left(
\mathbf{\Gamma}_i,\; \mathrm{OHE}(k_j^{gt})
\right),
\end{align}
where $k_j^{gt}$ denotes the ground-truth cardinality of pattern $j$, $\mathrm{OHE}(k_j^{gt})$ is the one-hot encoding of $k_j^{gt}$, and $\mathbf{C}_{i,k_j^{gt}} \in \mathbb{R}^{n_\mathrm{atom}}$ is the predicted node distribution conditioned on cardinality $k_j^{gt}$. 
The first term measures the alignment between predicted and ground-truth breakpoints, while the second term supervises the predicted cardinality.

The remaining question is how to map each query $i$ to the ground-truth pattern $j$ while keeping the permutation invariance.
We compute an optimal bipartite matching between the predicted and ground-truth patterns using the Hungarian algorithm, and optimize the model based on the resulting matched pairs. Specifically, we find permutation $\pi(\cdot)$ that minimizes the per-layer auxiliary loss,
\begin{align}
    \hat{\mathcal{L}}_\mathrm{fragment}=\min_{\pi} \sum_{i=1}^{n_\mathrm{query}}\hat{\mathcal{L}}_{\text{breakpoint}}^{i\pi(i)},
\end{align}
    where $\hat{\mathcal{L}}_{\text{breakpoint}}^{i\pi(i)}$ is the pairwise matching cost between query $i$ and ground-truth pattern with index $\pi(i)$.
Following DETR~\citep{carion2020endtoendobjectdetectiontransformers}, we further incorporate auxiliary decoding losses at the output of each transformer decoder layer to promote faster convergence and improved prediction accuracy. The final loss objective, $\mathcal{L}_\mathrm{fragment}$, is defined as a weighted sum of these auxiliary losses across all decoder layers, with decaying weights by a factor of $0.9$ with decreasing depth.

\textbf{Breakpoint Inference}.
At inference time, each query produces a breakpoint pattern by first selecting its predicted cardinality,
$
   \hat{k}_i = \arg\max \mathbf{\Gamma}_i.
$
Conditioned on this cardinality, we extract the corresponding node scores $\mathbf{C}_{i,\hat{k}_i} \in \mathbb{R}^{n_\mathrm{atom}}$ and construct the predicted breakpoint mask $\hat{\mathbf{b}}_i \in \{0,1\}^{n_\mathrm{atom}}$ by selecting the top-$\hat{k}_i$ nodes:
\begin{align}
    \hat{\mathbf{b}}_i = \mathrm{LinSATNet}(\mathbf{C}_{i,\hat{k}_i}, \hat{k}_i, 0).
\end{align}
Given breakpoints, fragments are further constructed as connected components after removing all predicted breakpoints. The details of GPU-friendly fragment construction during inference can be found in Appendix~\ref{sec:frag_inference}.

\subsection{Spectral Intensity Predictor}
\label{sec:intensity}
Existing two-stage models~\citep{Wang2025.05.28.656653,young2024fragnnet} compute embeddings for each fragment subgraph independently, leading to significant computational overhead due to redundant message passing over overlapping regions of the molecular graph. 
To address this inefficiency and fully leverage the expressive power of the Graphormer encoder, we propose an intensity predictor with an efficient strategy for subgraph pooling. Prior work in graph edit distance learning~\citep{wang2021combinatorial} shows that one can simply pool node embeddings to build subgraph features instead of rerunning message passing on every subgraph. This strategy, known as dynamic embedding, inspire the design of the intensity model.

Given a set of fragment masks $\mathcal{F}$ (where $|\mathcal{F}| = n_\mathrm{frag}$) a molecular graph, and the learned graph-level token $\mathbf{G}$, we construct fragment-level embeddings $\mathbf{H}$ using chemically relevant features (see Appendix~\ref{section:frag_embedding} for details), where each $\mathbf{H}_i$ corresponds to a fragment ${F}_i \in \mathcal{F}$.

We then refine these embeddings using a transformer decoder that performs masked attention over the full set of node embeddings $\mathbf{N}$:
\begin{align}
    \bar{\mathbf{H}} = \mathrm{TransformerDecoder}(\mathbf{H}, \mathbf{N}, \mathcal{F}),
\end{align}
where $\mathbf{N}$ serves as the memory and $\mathcal{F}$ defines binary attention masks that restrict each fragment to attend only to its corresponding nodes.
To further capture interactions between fragments, we apply a transformer encoder over the fragments:
\begin{align}
    \hat{\mathbf{H}} = \mathrm{TransformerEncoder}(\bar{\mathbf{H}}).
\end{align}
Finally, fragment intensities are predicted by passing the refined embeddings $\hat{\mathbf{H}}$ through an MLP that maps each fragment embedding to a 26 dimensional vector that represents 13 hydrogen shift states (losing or obtaining up to 6 hydrogen atoms) and both charge retention fragmentation
(CRF) and charge migration fragmentation (CMF). This setup is modified from~\cite{Wang2025.05.28.656653} with details in Appendix~\ref{section:inten_prediction_head}.

\subsection{Training Objective}\label{section:overall_objective}
The overall training objective consists of three components: the breakpoint prediction loss (discussed in Section~\ref{section:breakpoint}), a spectral distance loss, and an auxiliary ground-truth fragment spectral loss. We define the spectral distance loss as
\begin{align}
\mathcal{L}_\mathrm{inten} = \mathcal{D}(s^{gt}, \hat{s}), 
\label{eq:full_e2e_loss}
\end{align}
where $\mathcal{D}$ denotes the spectral cosine distance, $s^{gt}$ is the ground-truth spectrum, and $\hat{s}$ is the predicted spectrum. All spectral vectors are discretized into m/z bins with a width of 0.1 Dalton or 0.01 Dalton depending on the dataset specification.

To further stabilize training, we introduce an auxiliary loss $\hat{\mathcal{L}}_\mathrm{inten}$ with the same formulation as $\mathcal{L}_\mathrm{inten}$, but computed by feeding heuristically-labeled fragment patterns into the Intensity Predictor, similar to teacher forcing. This auxiliary objective helps stabilize the learning of the Intensity Predictor, particularly during the early stages of training. We validate its importance in Table~\ref{tab:ablation-magma}.

The full training objective is defined as
\begin{align}
\mathcal{L}_\mathrm{train} = w_\mathrm{magma}\cdot\left(w_\mathrm{inten}\cdot\hat{\mathcal{L}}_\mathrm{inten} + w_\mathrm{frag}\cdot\mathcal{L}_\mathrm{fragment}\right) + (1-w_\mathrm{magma})\cdot w_\mathrm{inten}\cdot\mathcal{L}_\mathrm{inten},
\label{eq:loss}
\end{align}
where $w_\mathrm{inten}$ and $w_\mathrm{frag}$ are fixed weighting hyperparameters, and $w_\mathrm{magma}$ starts from 1 for $n_\mathrm{magma}$ steps and and exponentially decays by a fixed rate $\lambda_\mathrm{magma}$ after each $n_\mathrm{decay}$ steps, where $n_\mathrm{magma}$, $n_\mathrm{decay}$, and $\lambda_\mathrm{decay}$ are the hyperparameters that control the number of warmup steps, the number of decay steps, and the decay rate, respectively. This design initializes fragment prediction using MAGMa heuristic~\citep{MAGMA} and stabilizes training in the early phase, while gradually shifting focus toward the real spectral prediction objective as training progresses.

Finally, we adopt a separate contrastive finetuning step following ICEBERG~\citep{Wang2025.05.28.656653} to further distinguish isomers, improving the downstream retrieval accuracy. We provide the details of contrastive finetuning in Appendix~\ref{section:contr_finetune}.

\section{Experiments}
\label{section:exp}
We evaluate \ours on established benchmarks for computational mass spectrometry, including NIST'20~\citep{nist_database} and MassSpecGym~\citep{bushuiev2024massspecgym}. We focus on two tasks: molecular retrieval and spectrum prediction (forward prediction). The experiments are designed to answer the following research questions: 
{(1)}~\textit{Does \ours improve the performance of mass spectrum prediction compared to two-stage fragment-prediction models and other state-of-the-art models?}
{(2)}~\textit{How accurate is \ours on an authentic assessment of identifying unknown compounds to support its application in real-world chemical and biological campaigns?}
{(3)}~\textit{Does \ours generate reasonable fragmentation patterns that align with the chemical intuition?}
{(4)}~\textit{Does \ours accelerate the inference pipeline with its single-stage integration?}
All experiments are conducted on a workstation with AMD 3995WX CPU, 4$\times$NVIDIA A5000 GPU, and 512GB RAM. We conduct experiments and answer these research questions quantitatively and qualitatively. 


\subsection{Datasets and Evaluation Metrics}
\textbf{NIST'20}.
We train our models on the NIST'20~\citep{nist_database} dataset with 530,640 high-energy collision-induced dissociation (HCD) spectra and 25,541 unique molecular structures. We evaluate on two different splits: (1) a random split that partitions data by InChI key and (2) a Murcko scaffold~\cite{bemis1996properties} split that clusters different molecular scaffolds and requires more generalization to out-of-distribution scaffolds. We include all peer methods reported in \citet{Wang2025.05.28.656653} for benchmarking purposes.

\textbf{MassSpecGym}.
We adopt the official MassSpecGym data split, which contains 231,104 tandem mass spectra (MS/MS) spanning 31,602 unique molecules. Importantly, this benchmark uses a Maximum Common Edge Subgraph (MCES)-based splitting strategy. The MCES protocol enforces substantial structural dissimilarity between training and testing molecules (edit distance $\geq 10$). As a result, it provides a more stringent evaluation of a model’s ability to generalize to previously unseen chemical space, rather than relying on memorization of known structural patterns.

\textbf{Evaluation Metrics}.
We evaluate \ours's capability in spectral simulation using top-$k$ retrieval accuracy, cosine similarity, and Jensen-Shannon similarity (interchangeable with spectral entropy similarity~\citep{li2021entropy}). Binned spectral vectors are used with 0.1 resolution on the NIST'20 dataset and 0.01 resolution on the MassSpecGym dataset to align with prior work on both benchmarks. 

\subsection{Results and Discussions}
\subsubsection{Spectral Prediction Similarity and Visualization}

\begin{figure*}[tb!]
    \centering

    \includegraphics[width=0.48\textwidth]{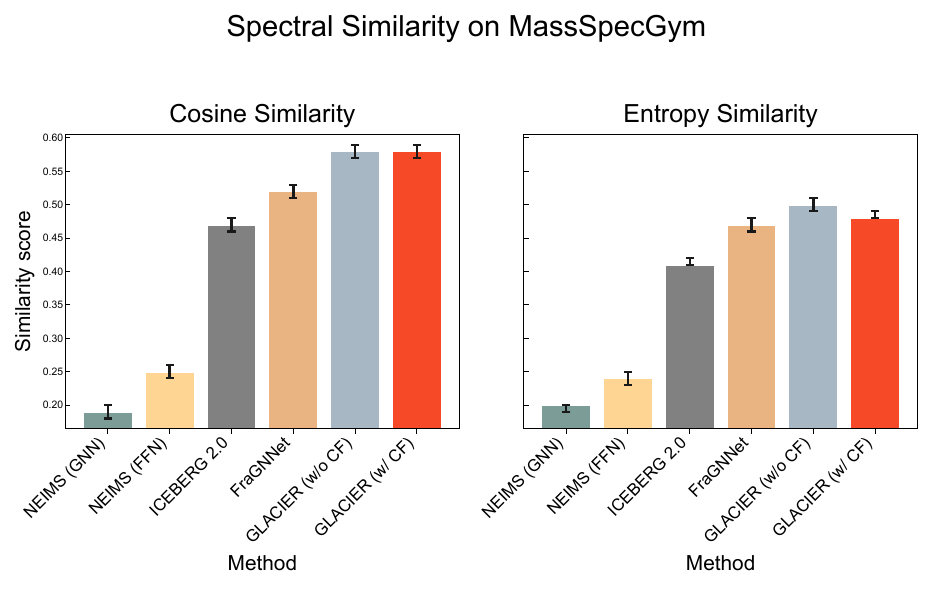}
    \hfill
    \includegraphics[width=0.48\textwidth]{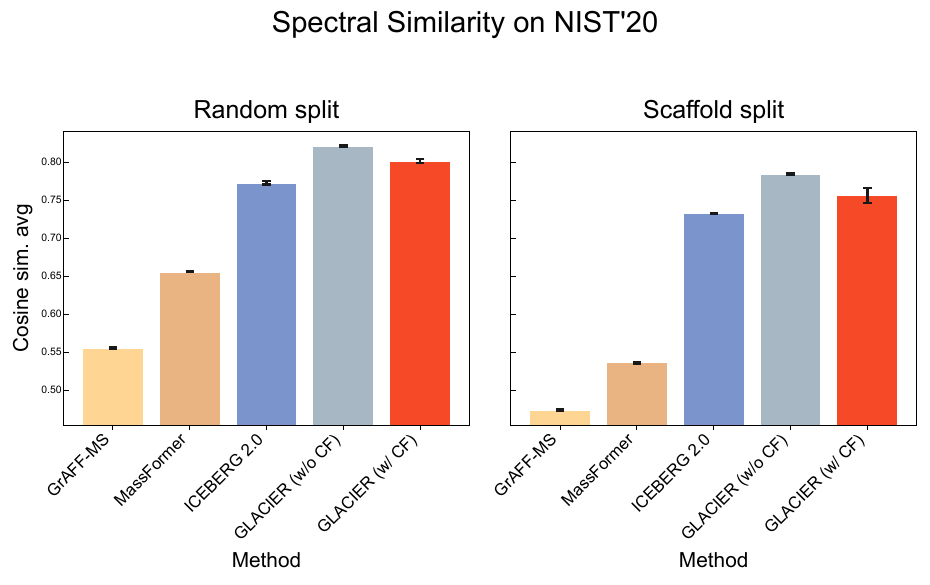}
    \vspace{-10pt}
    \caption{
    \textbf{Left}: Cosine and entropy similarity between experimental spectra and predictions on the MassSpecGym~\cite{bushuiev2024massspecgym} dataset.
    \textbf{Right}: Cosine similarity between experimental spectra and predictions on the NIST'20~\cite{nist_database} dataset with all adduct types. This plot does not include MolSpecFlow because it uses a different bin width.
    CF: contrastive finetuning.
    }
    
    \label{fig:cosine_sim_all}
\end{figure*}

We evaluate the prediction power of \ours by comparing the spectral similarity between the predicted spectrum and the experimental spectrum on the MassSpecGym and the NIST'20 datasets. 
As shown in Figure~\ref{fig:cosine_sim_all} (left), on MassSpecGym, \ours demonstrates better spectral similarity to ICEBERG.  
In Figure~\ref{fig:cosine_sim_all} (right), \ours outperforms all baselines on the NIST'20 dataset. Error bars are reported on three random seeds. Comparison with more baselines that only supports positive adduct types and the [M+H]\textsuperscript{+} adduct type along with the peak coverage statistics is reported in Appendix~\ref{section:spec_sim_extra}, where \ours still outperforms all prior methods. 
To evaluate MS/MS predictions qualitatively, we report a visualization of predicted spectra in Figure \ref{fig:spec} in the Appendix.

\subsubsection{Molecular Retrieval}
We evaluate the real-world applicability of MS/MS simulators and summarize the evaluation results in Table~\ref{table:retrieval_msg_mass} and Table~\ref{table:retrieval_NIST20_random}. \ours outperforms all baselines on MassSpecGym retrieval tasks, including the mass challenge and the formula bonus challenge, and the NIST'20 retrieval benchmark on both scaffold split and random split. We want to note that on MassSpecGym, the mass challenge is practically easier compared to the formula challenge, despite the name suggesting the opposite, because the number of candidates is always capped at 256 and the mass challenge introduces more easily distinguishable candidates that have different formulae. 
Results of retrieval accuracy of positive adduct types and [M+H]\textsuperscript{+} adducts on NIST'20 are included in Appendix~\ref{section:additional_retrieval} to compare with baselines that support fewer adducts, where \ours still outperforms prior methods.
\begin{table}[tb!]
    \centering
    \caption{Retrieval accuracy (mean with 99.9\% confidence intervals upon bootstrapping, 20,000 resamples) on the MassSpecGym dataset~\citep{bushuiev2024massspecgym}. \ours surpasses all baselines in retrieval accuracy.
    Note that MolSpecFlow~\citep{Wang2026.01.28.702438} does not have an open-source implementation and only reports on the mass challenge. It also uses a different bin width (0.1 Dalton) instead of the bin width adopted by MassSpecGym (0.01 Dalton); the comparison has to be made with extra care.}
    \vspace{-7pt}
    \resizebox{0.8\linewidth}{!}
    {
    \begin{tabular}{l|ccc}
    \toprule
    Model & Recall @ 1 & Recall @ 5 & Recall @ 20 \\
    \midrule
    
    \multicolumn{4}{c}{Mass Challenge: Sample up to 256 candidates with the same mass}\\
    \midrule
    
    NEIMS (FFN)~\citep{wei2019neims}
    & 8.44 (7.56-9.34)
    & 21.43 (20.10-22.79)
    & 38.57 (36.99-40.23)\\
    
    NEIMS (GNN)~\citep{zhu2020using}
    & 3.95 (3.37-4.62)
    & 11.92 (10.87-13.00)
    & 26.27 (24.83-27.82) \\
    
    FraGNNet~\citep{young2024fragnnet}
    & 46.64 (44.98-48.26)
    & 72.56 (71.18-74.00)
    & 83.58 (82.34-84.75) \\
    
    
    MolSpecFlow~\citep{Wang2026.01.28.702438}
    & 55.32 (52.97-56.87)
    & 79.98 (79.01-81.32)
    & 89.54 (88.32-90.76)\\
    
    ICEBERG~2.0~\citep{Wang2025.05.28.656653}
    & 63.95 (62.39-65.58)
    & 83.83 (82.56-84.99)
    & 92.56 (91.61-93.41) \\
        
    \midrule
    \textbf{{\ours}~(ours, w/o CF)} 
    & 69.69 (68.16-71.21)
    & 86.58 (85.41-87.69)
    & 93.34 (92.42-94.14) \\
    \textbf{{\ours}~(ours, w/ CF)}
    & \textbf{69.95 (68.36-71.50)}
    & \textbf{86.47 (85.30-87.56)}
    & \textbf{93.52 (92.62-94.31)}\\
    
    \midrule
    \multicolumn{4}{c}{Formula Challenge: Sample up to 256 candidates with the same chemical formula}\\
    \midrule
    
    NEIMS (FFN)~\citep{wei2019neims}
    & 7.62 (6.77-8.54)
    & 22.70 (21.32-24.12)
    & 44.12 (42.51-45.75)\\
    
    NEIMS (GNN)~\citep{zhu2020using}
    & 3.63 (3.05-4.29)
    & 13.55 (12.46-14.68)
    & 33.77 (32.26-35.37)\\
    
    FraGNNet~\citep{young2024fragnnet}
    & 31.93 (30.40-33.50)
    & 63.20 (61.64-64.76)
    & 82.70 (81.45-83.93)\\
    
    MARASON~\citep{wang2025neuralgraphmatchingimproves}
    & 34.03 (32.86-35.20)
    & 64.04 (62.77-65.19)
    & 85.39 (84.48-86.24)\\
    
    ICEBERG~2.0~\citep{Wang2025.05.28.656653}
    & 44.35 (42.75-46.02)
    & 73.57 (72.06-75.02)
    & 89.45 (88.44-90.43) \\
    
    \midrule
    \textbf{{\ours}~(ours, w/o CF)} 
    & 48.49 (46.88-50.15)
    & 76.56 (75.19-78.01)
    & 90.60 (89.62-91.57) \\
    
    \textbf{{\ours}~(ours, w/ CF)} 
    & \textbf{49.88 (48.25-51.49)}
    & \textbf{78.13 (76.78-79.47)}
    & \textbf{90.75 (89.75-91.70)} \\
    
    \bottomrule
    \end{tabular}
    }
    \vspace{-5pt}
    \label{table:retrieval_msg_mass}
\end{table}
\begin{table*}[tb!]
    \centering
    \caption{Retrieval accuracy (mean $\pm$ 95\% confidence interval on 3 random restarts) on NIST'20~\cite{nist_database} dataset on all adduct types. \ours surpasses all baselines in terms of retrieval accuracy. Results of all baselines are taken from \citet{Wang2025.05.28.656653}.} 
    \resizebox{\linewidth}{!}{
    \begin{tabular}{l|ccccccc}
    \toprule
        Accuracy @ Top-$k$ & 1 & 2 & 3  & 4 & 5 & 8 & 10 \\ \midrule
        \multicolumn{8}{c}{Random (InChI Key) Split}\\
        \midrule
        GrAFF-MS~\citep{murphy2023efficiently}
        & 0.165\textsubscript{$\pm$0.003}
        & 0.305\textsubscript{$\pm$0.003}
        & 0.406\textsubscript{$\pm$0.004}
        & 0.479\textsubscript{$\pm$0.005}
        & 0.535\textsubscript{$\pm$0.001}
        & 0.653\textsubscript{$\pm$0.004}
        & 0.708\textsubscript{$\pm$0.007} \\
        MassFormer~\citep{young2024massformer}
        & 0.212\textsubscript{$\pm$0.004}
        & 0.360\textsubscript{$\pm$0.005}
        & 0.465\textsubscript{$\pm$0.005}
        & 0.538\textsubscript{$\pm$0.004}
        & 0.595\textsubscript{$\pm$0.006}
        & 0.715\textsubscript{$\pm$0.003}
        & 0.765\textsubscript{$\pm$0.003} \\

        ICEBERG~2.0~\citep{Wang2025.05.28.656653}
        & 0.335\textsubscript{$\pm$0.000}
        & 0.549\textsubscript{$\pm$0.009}
        & 0.662\textsubscript{$\pm$0.004}
        & 0.733\textsubscript{$\pm$0.005}
        & 0.777\textsubscript{$\pm$0.004}
        & 0.858\textsubscript{$\pm$0.008}
        & 0.889\textsubscript{$\pm$0.005} \\

        \midrule
        \textbf{\ours (ours, w/o CF)}
        &
        0.390\textsubscript{$\pm$0.007}
        &	
        0.573\textsubscript{$\pm$0.003}
        &
        0.680\textsubscript{$\pm$0.007}
        & 
        0.741\textsubscript{$\pm$0.007}
        &
        0.785\textsubscript{$\pm$0.003}
        &
        0.861\textsubscript{$\pm$0.003}
        & 
        0.892\textsubscript{$\pm$0.004}
        \\
        \textbf{\ours (ours, w/ CF)} 
        & \textbf{0.525\textsubscript{$\pm$0.008}} 
        & \textbf{0.686\textsubscript{$\pm$0.006}} 
        & \textbf{0.774\textsubscript{$\pm$0.002}} 
        & \textbf{0.822\textsubscript{$\pm$0.002}} 
        & \textbf{0.857\textsubscript{$\pm$0.003}} 
        & \textbf{0.909\textsubscript{$\pm$0.005}} 
        & \textbf{0.925\textsubscript{$\pm$0.004}} \\
        \midrule

        \multicolumn{8}{c}{Murcko Scaffold Split}\\
        \midrule

        GrAFF-MS~\citep{murphy2023efficiently}
        & 0.140\textsubscript{$\pm$0.004}
        & 0.268\textsubscript{$\pm$0.005}
        & 0.364\textsubscript{$\pm$0.007}
        & 0.442\textsubscript{$\pm$0.005}
        & 0.506\textsubscript{$\pm$0.003}
        & 0.634\textsubscript{$\pm$0.003}
        & 0.698\textsubscript{$\pm$0.001} \\

        MassFormer~\citep{young2024massformer}
        & 0.179\textsubscript{$\pm$0.007}
        & 0.319\textsubscript{$\pm$0.004}
        & 0.420\textsubscript{$\pm$0.003}
        & 0.497\textsubscript{$\pm$0.002}
        & 0.559\textsubscript{$\pm$0.003}
        & 0.694\textsubscript{$\pm$0.005}
        & 0.757\textsubscript{$\pm$0.006} \\

        ICEBERG~2.0~\citep{Wang2025.05.28.656653}
        & 0.316\textsubscript{$\pm$0.007}
        & 0.524\textsubscript{$\pm$0.003}
        & 0.648\textsubscript{$\pm$0.005}
        & 0.720\textsubscript{$\pm$0.004}
        & 0.775\textsubscript{$\pm$0.005}
        & 0.866\textsubscript{$\pm$0.004}
        & 0.899\textsubscript{$\pm$0.008} \\

        \midrule
        \textbf{\ours (ours, w/o CF)} &
        0.384\textsubscript{$\pm$0.002}	&
        0.576\textsubscript{$\pm$0.007}	&
        0.692\textsubscript{$\pm$0.007}	&
        0.761\textsubscript{$\pm$0.003} &
        0.809\textsubscript{$\pm$0.007} &
	    \textbf{0.885\textsubscript{$\pm$0.003}} &
        \textbf{0.912\textsubscript{$\pm$0.003}} \\
        \textbf{\ours (ours, w/ CF)} 
        & \textbf{0.460\textsubscript{$\pm$0.004}} 
        & \textbf{0.636\textsubscript{$\pm$0.004}} 
        & \textbf{0.731\textsubscript{$\pm$0.009}} 
        & \textbf{0.788\textsubscript{$\pm$0.010}} 
        & \textbf{0.826\textsubscript{$\pm$0.009}} 
        & 
        \textbf{0.885\textsubscript{$\pm$0.007}}
        & 
        0.907\textsubscript{$\pm$0.007} \\
        \bottomrule
    \end{tabular}
    \vspace{-5pt}
    \label{table:retrieval_NIST20_random}
    }
\end{table*}

\subsubsection{Analysis of Breakpoint Prediction}
One open question is whether a single-stage model can predict reasonable fragments as well as two-stage models. We visualize a predicted breakpoint pattern in Figure~\ref{fig:breakpoints} in the Appendix. The model captures several chemically plausible breakpoint patterns, although some predicted breakpoints are not chemically reasonable. Importantly, these implausible fragments are subsequently suppressed by the spectral prediction module during intensity prediction. Table~\ref{tab:spec_sim_hplus} also demonstrates a comprehensive coverage of peaks quantitatively. These results suggest that our architecture effectively mirrors the object detection paradigm, in which an initial set of coarse predictions is generated and then refined by downstream modules to produce the final output.

\subsubsection{Inference Speed}
We evaluate the inference speed of \ours against the state-of-the-art two-stage model, 
\begin{wraptable}[7]{r}{0.43\textwidth}
\vspace{-15pt}
\centering
\caption{Inference speed comparison.}
\label{table:inference_speed}
\resizebox{\linewidth}{!}
{
\begin{tabular}{lcc}
\toprule
Model & ms/spec & spec/s/GPU \\
\midrule
ICEBERG~2.0~\citep{Wang2025.05.28.656653} 
& 10.8 & 46.4 \\
\textbf{\ours} & \textbf{1.3} & \textbf{368.7} \\
\midrule
{Speedup} & \multicolumn{2}{c}{{7.95$\times$}}\\
\bottomrule
\end{tabular}
}
\vspace{-8pt}
\end{wraptable}
ICEBERG~2.0~\citep{Wang2025.05.28.656653}. Specifically, we measure the average time required to complete spectral prediction for the molecular retrieval task on the scaffold split of the NIST'20 dataset (requiring prediction of 2,707,082 spectra in total, including all decoy structures), averaged over three random seeds for both models. We use 2$\times$NVIDIA A5000 GPUs and the results are summarized in Table~\ref{table:inference_speed}. \ours achieves a nearly 8-fold speedup compared to ICEBERG, highlighting its efficiency. This improvement is primarily attributed to our one-step, fully GPU-based fragment inference, which eliminates the need for CPU-intensive multi-step fragment enumeration. Speed comparison to MolSpecFlow~\citep{Wang2026.01.28.702438} is not possible because it is not open-sourced, and its paper does not report any speed statistics. 

\begin{table*}[tb!]
    \centering
    \caption{Ablation study of contrastive finetuning (CF) and its impact on Top-1 retrieval accuracy and spectral similarity. All metrics are reported in the same way with CI as in Table~\ref{table:retrieval_msg_mass} and Table~\ref{table:retrieval_NIST20_random}.}
    \label{tab:contrastive_ablation}
    \resizebox{\linewidth}{!}{
    \begin{tabular}{l|cccc|cccc}
        \toprule
        & \multicolumn{4}{c|}{MassSpecGym {(99.9\% CI)}} 
        & \multicolumn{4}{c}{NIST'20 {($\pm$95\% CI)}} \\
        \cmidrule(lr){2-5} \cmidrule(lr){6-9}
        Model 
        & Mass Top-1 
        & Formula Top-1 
        & Cos.\ Sim. 
        & J-S Sim.
        & Rand.\ Top-1 
        & Scaff.\ Top-1 
        & Rand.\ Cos. 
        & Scaff.\ Cos. \\
        \midrule
        ICEBERG~2.0 (w/ CF) 
        & 63.95\textsubscript{(62.39--65.58)}
        & 44.35\textsubscript{(42.75--46.02)}
        & 0.47\textsubscript{(0.46--0.48)}
        & 0.41\textsubscript{(0.41--0.42)}
        & 0.335\textsubscript{$\pm$0.000}
        & 0.316\textsubscript{$\pm$0.007}
        & 0.773\textsubscript{$\pm$0.002}
        & 0.733\textsubscript{$\pm$0.000} \\
        \midrule 
        
        \ours (w/o CF) 
        &69.69\textsubscript{(68.16--71.21)} 
        &48.49\textsubscript{(46.88--50.15)} 
        &\textbf{0.58\textsubscript{(0.57--0.59)}}
        &\textbf{0.50\textsubscript{(0.49--0.51)}}
        &0.390\textsubscript{$\pm$0.007}
        &0.384\textsubscript{$\pm$0.002}
        &\textbf{0.821\textsubscript{$\pm$0.001}}
        &\textbf{0.785\textsubscript{$\pm$0.002}}
        \\
        \ours (w/ CF) 
        &\textbf{69.95\textsubscript{(68.36--71.50)}} 
        &\textbf{49.88\textsubscript{(48.25--51.49)}}
        &\textbf{0.58\textsubscript{(0.57--0.59)}} 
        &0.48\textsubscript{(0.48--0.49)} 
        &\textbf{0.525\textsubscript{$\pm$0.008}} 
        &\textbf{0.460\textsubscript{$\pm$0.004}}
        &0.802\textsubscript{$\pm$0.002}
        &0.756\textsubscript{$\pm$0.010}\\
        \bottomrule
    \end{tabular}
    }
    \vspace{-15pt}
\end{table*}


\subsubsection{Ablation Studies}
\label{sec:ablation}
\textbf{MAGMa supervision weight decaying}. We study the impact of MAGMa heuristic and weight decaying in the loss function Eq.~(\ref{eq:loss}). Table~\ref{tab:ablation-magma} reports experiment results on the NIST'20 dataset 
\begin{wraptable}[6]{r}{0.4\textwidth}
    \vspace{-17pt}
    \caption{MAGMa supervision ablation.}
    \centering
    \resizebox{\linewidth}{!}
    {
    \begin{tabular}{cc|c}
    \toprule
    $\lambda_\mathrm{decay}$ & $w_\mathrm{magma}$ & Test Loss \\
       \midrule 
        0 & always $=0$ & Not Converged \\
        \textbf{0.9} & decays from 1 & \textbf{0.177} \\
       1 & always $=1$ & 0.243\\ 
       \bottomrule
    \end{tabular}
    }
    \label{tab:ablation-magma}
\end{wraptable}
under a random split with seed=1. If we rely on the intensity objective alone (first row), the model does not converge at all. If we always utilize MAGMa supervision and enforce teacher forcing (last row), the model converges but performs worse compared to the balanced weight-decaying strategy (second row). We let the model initialize for 2,500 steps before starting to decay $w_{\mathrm{magma}}$. Our explanation 
is that MAGMa-derived signals are particularly important during the early stages of training. As training progresses, the intensity decoder becomes capable of predicting low intensities for false positive fragments, allowing the model to rely more heavily on the full spectral prediction objective, the second term of Eq.~(\ref{eq:loss}).

\textbf{Alternative breakpoint prediction and end-to-end training objective}. We test an alternative fragment prediction strategy that directly predicts the atom-level subgraph mask, $\mathbf{m}$ by adapting GraphDETR~\citep{chen2026endtoendsubgraphdetectiongraphdetr} instead of predicting breakpoints as described in Section~\ref{section:breakpoint}. We discover that the performance is suboptimal compared with our breakpoint prediction approach, suggesting that direct fragment prediction is not a trivial task. We also implemented an end-to-end model with spectral similarity loss as the only objective and the model fails to converge. We summarize the results in Table~\ref{tab:ablation-fragment}. Our boundary-based prediction approach enforces subgraph connectivity and relaxes the precision needed to generate meaningful fragment candidates, ultimately leading to more stable training and improved downstream performance. 

\begin{wraptable}[6]{r}{0.3\textwidth}
    \vspace{-19pt}
    \caption{Fragment Prediction ablation.}
    \centering
    \resizebox{\linewidth}{!}
    {
    \begin{tabular}{c|c}
    \toprule
    Prediction strategy & Test Loss \\
       \midrule 
        end-to-end & not converge \\
        GraphDETR & 0.477 \\
       Breakpoint & 0.418\\ 
       \bottomrule
    \end{tabular}
    }
    \label{tab:ablation-fragment}
\end{wraptable}

\textbf{Contrastive finetuning}. Contrastive finetuning of spectral prediction to improve retrieval performance was introduced by ICEBERG~2.0~\cite{Wang2025.05.28.656653}, which samples negative structures from PubChem and by random mutation, and separates them from the positive structure. The technical details are discussed in Appendix~\ref{section:contr_finetune}. As shown in Table~\ref{tab:contrastive_ablation}, we find contrastive finetuning is still an important technique for \ours, where it introduces a slight negative impact on spectral accuracy but significant improvement to retrieval accuracy. The base version of \ours without contrastive finetuning already outperforms ICEBERG. This technique seems to benefit more for in-distribution data, as the improvement observed is the most significant on NIST'20 random split and the least significant on MassSpecGym. 
There are several hyperparameters for contrastive finetuning. We further show that \ours works better with ``harsher'' contrastive conditions compared to that used by ICEBERG, as shown in Table~\ref{tab:contrastive_random_ablation}, \ref{tab:contrastive_scaffold_ablation}, \ref{tab:contrastive_msg_mass_ablation}, and \ref{tab:contrastive_msg_formula_ablation} in the Appendix. However, the harsher the contrastive condition, the lower the spectral similarity and peak coverage are, as summarized in Table~\ref{tab:CF_ablation_spec_sim}. This can be reasoned as the tradeoff between spectral similarity and distinguishability. As the CF condition gets harsher, the model will mitigate the impact of non-essential noise peaks from the spectra, resulting in lower spectral similarity and higher distinguishability. We acknowledge that depending on the method of sampling decoys, CF could present a risk of pathological shortcut-learning~\citep{liu2026massspecgymwilduncoveringcorrecting}, though that is not empirically the case in this work (Table~\ref{tab:contrastive_filter}; Appendix~\ref{section:contr_finetune}). 

\textbf{Bin width}. We study the effect of bin width resolution during training and evaluation on the MassSpecGym dataset. As shown in Appendix~\ref{section:bin_width_ablation}, we find that having higher bin-width resolution improves retrieval performance of \ours on both the formula challenge and the mass challenge. This demonstrates the effectiveness of the fragment-based prediction approach in high-resolution spectral prediction task due to its capability of predicting exact m/z ratios using fragment mass.

\textbf{Alternative graph encoder}
We study the influence of different graph encoder backbones on the prediction power of \ours on the MassSpecGym dataset. We tested 3 graph encoder variants: Graphormer, NeuralWalker~\citep{chen2024learninglongrangedependencies} (which is used in GraphDETR), and vanilla GNN.    
\begin{wraptable}[7]{l}{0.34\textwidth}
    \vspace{-20pt}
    \caption{Graph encoder ablation.}
    \centering
    \resizebox{\linewidth}{!}
    {
    \begin{tabular}{c|c}
    \toprule
    Graph Encoder & Test Loss \\
       \midrule 
        GNN & 0.433 \\
        NeuralWalker & 0.423 \\
       Graphormer & 0.418\\ 
       \bottomrule
    \end{tabular}
    }
    \label{tab:ablation-graph-encoder}
\end{wraptable}
As shown in Table~\ref{tab:ablation-graph-encoder}, choosing Graphormer as the graph encoder backbone results in the best test loss, supporting our choice of adopting it as the default graph encoder backbone used in \ours.
\textbf{Alternative fragment encoding initialization.}
In order to make sure that \ours does not heavily rely on predefined heuristic and learn useful representation, we added ablations using two alternatives: (1) Transformer-decoder pooling without descriptors and (2) Transformer-decoder pooling without graph tokens, with results summarized in Table~\ref{tab:ablation-graph-encoder}. The results show that removing external descriptors does not negatively affect performance. This demonstrates \ours’s ability to learn effective atom-level and graph-level representation without relying on external descriptors.
\begin{wraptable}[5]{r}{0.34\textwidth}
    \vspace{-25pt}
    \caption{Fragment token ablation.}
    \centering
    \resizebox{\linewidth}{!}
    {
    \begin{tabular}{c|c}
    \toprule
    Initialization method & Test Loss \\
       \midrule 
        graph token & 0.411 \\
        graph token+descriptor & 0.418 \\
       descriptor & 0.421\\ 
       \bottomrule
    \end{tabular}
    }
    \label{tab:ablation-graph-encoder}
\end{wraptable}
\section{Discussion and Outlook}
\label{section:conclusion}
In this paper, we establish the conceptual connection between MS/MS prediction in molecular machine learning and object detection in computer vision.
We design \ours, a transformer-based single-stage MS/MS prediction model that supports multiple bond-breaking events and accurate intensity prediction. 
\ours establishes new state-of-the-art performance with better retrieval accuracy in downstream applications and a significant boost in inference throughput. 
We envision that the development of \ours will open up a new space of model design for single-stage MS/MS prediction networks and raise attention from the broader machine learning community, whereby faster and more accurate models will shed light on the long-lasting challenge of structural elucidation, enabling critical science areas including clinical metabolomics, precise medicine, and drug discovery.

\textbf{Limitations and future work}. As the first model for one-stage multi-breakpoint MS/MS predictions, \ours poses several limitations to be addressed in future work. First, though being one-stage, \ours is not a fully end-to-end model. The fragment generation step truncates the gradient, and we use MAGMa heuristic to supervise fragment generation. An end-to-end trainable one-stage model in future work might further improve the performance. Second, \ours does not handle complex rearrangements that involve changes in the heavy atom connectivity. Future work could focus on bond forming and breaking as the fragment generation step. 

\textbf{Broader impact}. Tools developed in this paper for structural elucidation have broad potential for positive societal impact across chemical and biomedical sciences, with minimal foreseeable risk.

\begin{ack}
This work was supported by DSO National Laboratories in Singapore, the MIT Generative AI Impact Consortium (MGAIC), and the National Institute of General Medical Sciences of the National Institutes of Health (K99GM166738). The authors thank Shitong Luo and Mrunali Manjrekar for discussions and Hongxuan Liu for suggestions on parallel inference. 
\end{ack}

\bibliographystyle{plainnat}
\bibliography{reference}







\newpage
\appendix

\section*{Technical Appendices and Supplementary Materials}
\section{Graphormer Node and Edge Features} \label{section:features}
To construct more expressive node and edge representations for the molecular graph, we incorporate chemically informative features derived from RDKit and use heavy atom scaffold as the graph. Because our model shares embeddings across multiple subgraph fragments from the same molecule, we can compute these features once at the molecule level and reuse them across fragments. In contrast, prior work~\citep{Wang2025.05.28.656653} generates separate embeddings for each fragment, which prevents efficient integration of such global, RDKit-derived features. We summarize the resulting atom and edge features in Table~\ref{tab:node_features} and Table~\ref{tab:edge_features}.
\begin{table}[th]
\caption{Node feature construction for each atom. Here $C = \texttt{MAX\_ABS\_FORMAL\_CHARGE}$.}
\centering
\small
\resizebox{\textwidth}{!}{
\begin{tabular}{lll}
\toprule
\textbf{Feature Name} & \textbf{Description} & \textbf{Dimension} \\
\midrule
Atom Type (One-hot) & One-hot over \texttt{VALID\_ATOM\_NUM} & $|\texttt{VALID\_ATOM\_NUM}|$ \\
Atom Type (Other) & Indicator for atoms not in valid set & $1$ \\

Degree (One-hot) & One-hot over $0 \ldots \texttt{MAX\_COMMON\_DEGREE}$ (excl. H) & $\texttt{MAX\_COMMON\_DEGREE}+1$ \\
Degree (Overflow) & Indicator for degree $>$ max & $1$ \\

Formal Charge (One-hot) & One-hot over $[-C, \ldots, C]$ & $2C+1$ \\
Formal Charge (Overflow) & Indicator for $|q| > C$ & $1$ \\

Hybridization (One-hot) & One-hot over \texttt{COMMON\_HYBRIDIZATION} & $|\texttt{COMMON\_HYBRIDIZATION}|$ \\
Hybridization (Other) & Indicator for unseen hybridization & $1$ \\

Aromaticity & Atom is aromatic & $1$ \\

Hydrogen Count (One-hot) & One-hot over $0 \ldots \texttt{COMMON\_MAX\_HYDROGEN\_COUNTS}$ & $\texttt{COMMON\_MAX\_HYDROGEN\_COUNTS}+1$ \\
Hydrogen Count (Overflow) & Indicator for count $>$ max & $1$ \\

Ring Membership & $[\text{not in ring}, \text{in ring}]$ & $2$ \\

Ring Size (Multi-hot) & For each size in \texttt{COMMON\_RING\_SIZES}: $[\text{in}, \text{not in}]$ & $2 \times |\texttt{COMMON\_RING\_SIZES}|$ \\

Fused Ring & $[\text{not fused}, \text{fused}]$ & $2$ \\

Chirality (One-hot) & One-hot over \texttt{COMMON\_CHIRALITY} & $|\texttt{COMMON\_CHIRALITY}|$ \\

Element Group Embedding & Element group representation (optional) & $\texttt{ELEMENT\_GROUP\_DIM}$ \\

Gasteiger Charge & Continuous partial charge & $1$ \\
\bottomrule
\end{tabular}
}
\label{tab:node_features}
\end{table}
\begin{table}[th]
\caption{Edge (bond) feature construction. Fused rings are defined as bonds belonging to $\geq 2$ rings.}
\centering
\small
\begin{tabular}{lll}
\toprule
\textbf{Feature Name} & \textbf{Description} & \textbf{Dimension} \\
\midrule
Bond Type (One-hot) & One-hot over \texttt{COMMON\_BOND\_TYPES} & $|\texttt{COMMON\_BOND\_TYPES}|$ \\
Bond Type (Other) & Indicator for unseen bond types & $1$ \\

Conjugation & $[\text{not conjugated}, \text{conjugated}]$ & $2$ \\

Ring Membership & $[\text{not in ring}, \text{in ring}]$ & $2$ \\

Ring Size (Multi-hot) & For each size in \texttt{COMMON\_RING\_SIZES}: $[\text{in}, \text{not in}]$ & $2 \times |\texttt{COMMON\_RING\_SIZES}|$ \\

Fused Ring & $[\text{not fused}, \text{fused}]$ & $2$ \\

Stereo (One-hot) & One-hot over \texttt{VALID\_BOND\_STEREO} & $|\texttt{VALID\_BOND\_STEREO}|$ \\
Stereo (Other) & Indicator for unseen stereo types & $1$ \\

\bottomrule
\end{tabular}
\label{tab:edge_features}
\end{table}

\section{Fragment Pattern Construction in Inference}
\label{sec:frag_inference}
Once the breakpoints are predicted, the next step is to generate the fragments. Let $\mathbf{A} \in \{0,1\}^{n_\mathrm{atom} \times n_\mathrm{atom}}$ denote the adjacency matrix of a graph and let 
$
\mathbf{m}^{(l)} \in \{0,1\}^{n_\mathrm{atom}}$ denote the $l$-th breakpoint mask for a given molecular graph. 

\textbf{Reachability.}
For each mask $\mathbf{m}^{(l)}$, we construct a masked adjacency matrix:
\begin{align}
    \mathbf{A}^{(l)} = \mathbf{A} \odot \big(\mathbf{m}^{(l)} (\mathbf{m}^{(l)})^\top \big),
\end{align}
which restricts edges to those entirely within the subgraph.

We compute the reachability matrix $\mathbf{R}^{(l)} \in \{0,1\}^{n_\mathrm{atom} \times n_\mathrm{atom}}$ as the transitive closure of $\mathbf{A}^{(l)}$:
\begin{align}
\mathbf{R}^{(l)}_{ij} = \mathbb{I}\big( \text{there exists a path from } i \text{ to } j \text{ within } \mathbf{A}^{(l)} \big).
\end{align}
\textbf{Component extraction.}
Each row $\mathbf{R}^{(l)}_i$ encodes the set of nodes reachable from node $i$, corresponding to a connected component within the masked subgraph. 
We treat each row as a candidate fragment pattern:
$
\mathbf{p}^{(l,i)} = \mathbf{R}^{(l)}_i \in \{0,1\}^{n_\mathrm{atom}}.
$
Since multiple nodes within the same connected component yield identical rows, we remove duplicate patterns to get the fragment set $\mathcal{F}$:
\begin{align}
    \mathcal{F} = \mathrm{Unique}\big( \{ \mathbf{p}^{(l,i)} \}_{l,i} \big).
\end{align}
We discard trivial patterns with no active nodes, and $\mathbf{p}$ is kept only if $\sum_j p_j > 0$.
In the degenerate case where $| \mathcal{F}| = 0$, we insert a dummy pattern corresponding to valid nodes.

\section{Fragment Embedding for Intensity Prediction}
\label{section:frag_embedding}

We construct fragment-level embeddings by incorporating chemically meaningful structural signals. We first generate fragment formula vectors, encoding chemical formula as a multi-hot vector for each fragment. In parallel, we compute fragment-specific statistics, including the number of hydrogens within each fragment, the maximum allowable hydrogen addition and removal (i.e., hydrogen shift, bounded by the number of broken bonds), and the number of broken bonds, which we define as edges crossing the fragment boundary. We derive these quantities using adjacency matrices and fragment membership masks. We then compare each fragment form vector with the corresponding root (molecule-level) form vector to obtain difference features. We pass both the fragment representations and their differences through a shared embedding network. Additionally, we discretize the number of broken bonds using a clamped one-hot encoding. Finally, we concatenate the expanded root token embeddings, encoded fragment features, difference encodings, and broken-bond features, and map the result through a learned projection layer to produce the final token representations. In particular, we can define the fragment embedding token $\mathbf{t}_k$ as follows of fragment $k$:
\begin{align}
\mathbf{t}_{k} = \mathrm{MLP} \Big(
\big[
\mathbf{G} \;\|\;
\psi(\mathbf{f}_{k}) \;\|\;
\psi(\mathbf{r}^{\text{form}} - \mathbf{f}_{k}) \;\|\;
\mathrm{OHE}\big(n^{\text{broken}}_{k}\big)
\big]
\Big),
\end{align}
where $\mathbf{G}$ is the embedding of the root molecule, $\psi$ is the absolute-sine embedder for formula embedding, $\mathbf{r}^{\text{form}}$ is the formula vector of the root molecule, $\mathbf{f}_{k}$ is the formula vector of fragment $k$, and  $n^{\text{broken}}_{k}$ is the number of broken bonds required when generating fragment $k$.
\section{Intensity Prediction Head}
ICEBERG predicts unnormalized intensity weights at each possible hydrogen shift alongside an attention weight  to determine how heavily to weight each prediction for its specified hydrogen shift for all fragment embeddings. The attention weight is calculated by taking a softmax over all prediction indices that fall into the same intensity bin. The final binned intensity is calculated as the weighted sum over all predictions that fall within this mass bin followed by a sigmoid activation function. The final intensity is calculated as the weighted sum over all predictions that fall within this mass bin followed by a sigmoid activation function. The unbinned intensity of is reassigned back to each fragment with the predicted attention score that and the total intensity falls into a given bin.

This setting is problematic because it mixes up the assignment score before and after activation and will lead to inaccurate results if we want to adapt the predicted fragment intensity into different bin-width setting. To address this, we remove the attention weight component and predict the fragment intensity as follows: 
\begin{align}
y^{(i)}_\delta = \sigma\left(
\mathrm{MLP}_{\mathrm{inten}}
(\hat{\mathbf{H}}_i
)_\delta\right),
\end{align}
where $\sigma$ is the Sigmoid activation function, $\delta$ is the hydrogen shift count, and $y^{(i)}_\delta$ is the predicted unbinned intensity of the fragment with hydrogen shift $\delta$.
The binned intensity $\hat{y}_m$ at m/z value $m$ is calculated as the sum of all unbinned intensity predictions that fall within a given mass bin: 

\begin{equation}
\hat{y}_m = \sum_i \sum_\delta
y^{(i)}_\delta
\, \mathbb{I}[M(i,\delta) = m],
\end{equation}
where $M(i,\delta)$ is the m/z value of fragment $i$ with hydrogen shift $\delta$.
This formulation enables the model to adapt to different mass bin widths at inference time, even when trained using a fixed binning resolution.

\label{section:inten_prediction_head}
\section{Contrastive Finetuning Details}
\label{section:contr_finetune}
We adapt the formulation of the contrastive finetuning objective from~\citet{Wang2025.05.28.656653}.Of our sampled negative structures, 50\% share the same
chemical formula (from PubChem) and the remaining 50\% are generated by single-step graph mutations~\citep{tripp2023genetic}. During training, \ours predict spectra for one positive and $k$ negative
structures simultaneously, yielding the following pairwise distance matrix:
\begin{align}
\mathbf{D} \in \mathbb{R}^{(k+1)\times (k+1)}, \quad 
\mathbf{D}_{ij} = d_i - d_j, \quad i,j \in \{1, \dots, k+1\},
\end{align}
where $d_i$ is the entropy distance between the predicted spectrum for
structure $i$ and the ground-truth spectrum, and $d_1$ is the positive sample. We compute $\bar{\mathbf{D}} = \mathrm{Sinkhorn}(-\mathbf{D}, \tau)$ and $\bar{d_1} \in [0, 1]$ is the top-left entry of $\bar{\mathbf{D}}$. The temperature hyperparameter $\tau$ is set to $0.5$. We define our contrastive loss term $\mathcal{L}_\mathrm{contr}$ as follows:
\begin{align}  \mathcal{L}_\mathrm{contr}=\mathrm{ReLU}(-\log (\bar{d_1} + \delta)),
\end{align}
where the offset hyperparameter $\delta$ prevents overfitting by discouraging further maximization when $\bar{d_1}$ is already large.
The overall contrastive finetuning objective is defined as follows:
\begin{align}
\mathcal{L}^\mathrm{train}_\mathrm{contr} = w_\mathrm{magma}\cdot w_\mathrm{frag}\cdot\mathcal{L}_\mathrm{fragment} + w_\mathrm{inten}\cdot\mathcal{L}_\mathrm{inten}+w_\mathrm{contr}\cdot\mathcal{L}_\mathrm{contr},
\end{align}
where $w_\mathrm{magma}, w_\mathrm{frag}, \mathcal{L}_\mathrm{fragment}, w_\mathrm{inten},$ and $\mathcal{L}_\mathrm{inten}$ are described in Section~\ref{section:overall_objective}. Note that we remove the magma inten loss $\hat{\mathcal{L}}_\mathrm{inten}$ since the intensity predictor is pretrained. Similar to the pretraining stage, we also remove the fragmentation prediction objective $\mathcal{L}_\mathrm{fragment}$ and the overall contrastive finetuning objective becomes 
\begin{align}
   \mathcal{L}^\mathrm{val}_\mathrm{contr} = w_\mathrm{inten}\cdot\mathcal{L}_\mathrm{inten}+w_\mathrm{contr}\cdot\mathcal{L}_\mathrm{contr}.
\end{align}

We study the impact of the performance of contrastive finetuning hyperparameters on NIST'20 random split and scaffold split and on the MassSpecGym dataset. Results are summarized in Table~\ref{tab:contrastive_random_ablation}, Table~\ref{tab:contrastive_scaffold_ablation}, Table~\ref{tab:contrastive_msg_mass_ablation}, and Table~\ref{tab:contrastive_msg_formula_ablation}. We find out that using a ``harsher'' contrastive finetuning condition (lower $\delta$ and higher $d$) results in significant performance boost on random split while its impact on scaffold split is relatively limited. Our explanation is that contrastive finetuning is more effective for learning on in-distribution data where the model can learn to differentiate between similar molecular structures, while having distinct structures across training and validation sets makes it difficult for the model to separate structures it has not seen before.

\begin{table}[ht]
\centering
\caption{Top-$k$ accuracy comparison between three contrastive finetuning (CF) conditions on NIST'20 random split.}
\resizebox{\textwidth}{!}{
\begin{tabular}{l|cccccccccccc}
\toprule
 & Top-1 & Top-2 & Top-3 & Top-4 & Top-5 & Top-8  & Top-10 \\
\midrule
\ours ($\delta=0.4, d=5$ ) & \textbf{0.525\textsubscript{$\pm$0.008}} & \textbf{0.686\textsubscript{$\pm$0.006}} & \textbf{0.774\textsubscript{$\pm$0.002}} & 
\textbf{0.822\textsubscript{$\pm$0.002}} & 
\textbf{0.857\textsubscript{$\pm$0.003}} & 
\textbf{0.909\textsubscript{$\pm$0.005}} &		
0.925\textsubscript{$\pm$0.004} \\

\ours ($\delta=0.5, d=3$ ) &
0.498\textsubscript{$\pm$0.008} & 0.667\textsubscript{$\pm$0.010} & 
0.754\textsubscript{$\pm$0.009} & 0.808\textsubscript{$\pm$0.006} &
0.845\textsubscript{$\pm$0.005} & 	
0.907\textsubscript{$\pm$0.008} &	
\textbf{0.929\textsubscript{$\pm$0.007}} \\

\ours ($\delta=0.6, d=3$ ) & 0.486\textsubscript{$\pm$0.007} & 0.657\textsubscript{$\pm$0.004} &  0.748\textsubscript{$\pm$0.004} & 0.802\textsubscript{$\pm$0.003} & 0.840\textsubscript{$\pm$0.004} & 
0.906\textsubscript{$\pm$0.003} & \textbf{0.929\textsubscript{$\pm$0.005}}\\
\bottomrule
\end{tabular}
}
\label{tab:contrastive_random_ablation}
\end{table}

\begin{table}[h]
\centering
\caption{Top-$k$ accuracy comparison between three contrastive finetuning (CF) conditions on NIST'20 scaffold split.}
\resizebox{\textwidth}{!}{
\begin{tabular}{c|cccccccccccc}
\toprule
 & Top-1 & Top-2 & Top-3 & Top-4 & Top-5 & Top-8  & Top-10 \\
\midrule
\ours ($\delta=0.4, d=5$ ) &
\textbf{0.460\textsubscript{$\pm$0.004}} & \textbf{0.636\textsubscript{$\pm$0.004}} &
0.731\textsubscript{$\pm$0.009} &
0.788\textsubscript{$\pm$0.010} & 
0.826\textsubscript{$\pm$0.009} & 
0.885\textsubscript{$\pm$0.007} &
0.907\textsubscript{$\pm$0.007} \\

\ours ($\delta=0.5, d=3$ ) & 0.448\textsubscript{$\pm$0.004} & 0.633\textsubscript{$\pm$0.010} & \textbf{0.736\textsubscript{$\pm$0.008}} & 
\textbf{0.798\textsubscript{$\pm$0.010}} & 
\textbf{0.836\textsubscript{$\pm$0.006}} &
0.899\textsubscript{$\pm$0.004} & 
0.923\textsubscript{$\pm$0.001} \\
\ours ($\delta=0.6, d=3$ ) & 0.441\textsubscript{$\pm$0.006} & 0.625\textsubscript{$\pm$0.011} & 0.729\textsubscript{$\pm$0.012} & 0.793\textsubscript{$\pm$0.006} & 0.834\textsubscript{$\pm$0.007} & \textbf{0.901\textsubscript{$\pm$0.003}} & \textbf{0.926\textsubscript{$\pm$0.004}}\\
\bottomrule
\end{tabular}
}
\label{tab:contrastive_scaffold_ablation}
\end{table}

\begin{table}[h]
\centering
\caption{Top-$k$ accuracy comparison between three contrastive finetuning (CF) conditions on the MassSpecGym (mass challenge).}
\resizebox{\textwidth}{!}{
\begin{tabular}{c|cccccccccccc}
\toprule
 & Top-1 & Top-2 & Top-3 & Top-4 & Top-5 & Top-8  & Top-10 \\
\midrule
\ours ($\delta=0.4, d=5$ ) & 0.681 & 0.784 & 	0.823 & 0.847 & 0.862 & 0.890 & 0.901 \\

\ours ($\delta=0.5, d=3$ ) & 0.698 & \textbf{0.792} & 0.828 & 0.850 & 0.864 & \textbf{0.895} & \textbf{0.905} \\ 

\ours ($\delta=0.6, d=3$ ) & \textbf{0.699} & \textbf{0.792} & \textbf{0.832} & \textbf{0.853} & \textbf{0.865} & \textbf{0.895} & \textbf{0.905} \\
\bottomrule
\end{tabular}
}
\label{tab:contrastive_msg_mass_ablation}
\end{table}

\begin{table}[h]
\centering
\caption{Top-$k$ accuracy comparison between three contrastive finetuning (CF) conditions on the MassSpecGym (formula challenge).}
\resizebox{\textwidth}{!}{
\begin{tabular}{c|cccccccccccc}
\toprule
 & Top-1 & Top-2 & Top-3 & Top-4 & Top-5 & Top-8  & Top-10 \\
\midrule
\ours ($\delta=0.4, d=5$ ) & 0.490 & 0.624 & 0.692 & 0.738 & 0.770 & 0.827 & 0.849\\

\ours ($\delta=0.5, d=3$ ) & 0.490 & 0.627 & 	\textbf{0.700} & 0.745 & 0.775 & 0.835 & \textbf{0.855} \\ 

\ours ($\delta=0.6, d=3$ ) & \textbf{0.499} & \textbf{0.633} & 0.699 & \textbf{0.749} & \textbf{0.781} & \textbf{0.836}	& 0.853 \\
\bottomrule
\end{tabular}
}
\label{tab:contrastive_msg_formula_ablation}
\end{table}
Further, we do not currently believe that contrastive finetuning is a source of ``data leakage''--- the information provided by contrastive finetuning is that PubChem structure A should not have a spectrum that looks similar to structure B, which is in the training set. It might or might not be true in reality. As a standard approach, it does not seem to leak any information because all synthetic structures are used as negative samples in contrastive finetuning. To verify this idea, we conduct another ablation study testing whether filtering out all retrieval candidate structures in retrieval during contrastive finetuning training affects the overall performance, with results summarized in Table~\ref{tab:contrastive_filter}. \ours performs similarly with or without candidate structure filtering, suggesting that the resulting performance boost is not due to data leakage but the enhancement of model's ability to differentiate structures.  Nevertheless, we emphasize that even in the absence of any contrastive fine-tuning, GLACIER achieves state of the art retrieval accuracy for both MassSpecGym and NIST'20.

\begin{table}[h!]
\centering
\caption{Top-$k$ accuracy comparison between models on random split with the same CF condition ($\delta=0.5, d=3$) and random seed ($1$) on the NIST'20 dataset. \ours performs similarly across the two conditions, suggesting that the performance boost of contrastive finetuning is not the results of data leakage arising from inclusion of exactly matching decoys.}
\resizebox{\textwidth}{!}{
\begin{tabular}{c|cccccccccccc }
\toprule
 & Top-1 & Top-2 & Top-3 & Top-4 & Top-5 & Top-8  & Top-10 \\
\midrule
\ours (w/o decoy filtering) &
\textbf{0.510} & \textbf{0.677} & \textbf{0.761} & 0.814 & \textbf{0.849} & 0.905 & 0.928 \\
\ours (with decoy filtering) & 0.505 & 0.674 & 0.759 & \textbf{0.815} & 0.848 & \textbf{0.912} & \textbf{0.934} \\

\bottomrule
\end{tabular}
}
\label{tab:contrastive_filter}
\end{table}

\begin{table}[h!]
\caption{Ablation studies on the effect of different CF condition on spectral similarities and coverage on the NIST'20 dataset.}
\label{tab:CF_ablation_spec_sim}
\centering
{
\begin{tabular}{lccc}
\toprule
\multirow{2}{*}{Method} & \multicolumn{3}{c}{\textbf{Positive+negative mode, Random split}} \\
 & Cosine sim. ($\uparrow$) & Entropy sim. ($\uparrow$) & Coverage ($\uparrow$) \\
\midrule
\ours ($\delta=0.4, d=5$ ) & 0.802 $\pm$ 0.002 & 0.736 $\pm$ 0.004 & 0.843 $\pm$ 0.003 \\
\ours ($\delta=0.5, d=3$ ) & 0.820 $\pm$ 0.002 & 0.760 $\pm$ 0.002 & 0.862 $\pm$ 0.002\\
\ours ($\delta=0.6, d=3$ ) & \textbf{0.824 $\pm$ 0.000} & \textbf{0.767 $\pm$ 0.001} & \textbf{0.868 $\pm$ 0.000}\\
\midrule
\midrule
\multirow{2}{*}{Method} & \multicolumn{3}{c}{\textbf{Positive+negative mode, Scaffold split}} \\
 & Cosine sim. ($\uparrow$) & Entropy sim. ($\uparrow$) & Coverage ($\uparrow$) \\
\midrule
\ours ($\delta=0.4, d=5$ ) & 0.756 $\pm$ 0.010 & 0.687 $\pm$ 0.009 & 0.826 $\pm$ 0.008 \\
\ours ($\delta=0.5, d=3$ ) & 0.778 $\pm$ 0.002 & 0.717 $\pm$ 0.002 & 0.852 $\pm$ 0.003\\
\ours ($\delta=0.6, d=3$ ) & \textbf{0.787 $\pm$ 0.000} & \textbf{0.730 $\pm$ 0.001} & \textbf{0.862 $\pm$ 0.001}\\
\bottomrule
\end{tabular}
}
\end{table}
\clearpage

\section{Ablation Studies on Bin Width Resolution}
\label{section:bin_width_ablation}
We conduct ablation studies on the effect of retrieval accuracy on models trained on bin width of 0.1 Dalton or 0.01 Dalton and evaluate using 0.1 Dalton and 0.01 Dalton on the MassSpecGym formula and mass challenges, and summarize the result in Table~\ref{table:bin_width_ablation}. \ours performs better when tested on bin width set to 0.01 Dalton, demonstrating the power of fragment-based approach on high-resolution spectral prediction. 

\begin{table}[h]
\caption{Top 1 accuracy of \ours trained/tested on the MassSpecGym dataset with bin width set to 0.1 Dalton or 0.01 Dalton. Rows correspond to the training bin width and columns correspond to the testing bin width. \ours performs better using 0.01 Dalton as the testing bin widths.}
    \centering
    \begin{tabular}{c|cc|cc}
        \toprule
        & \multicolumn{2}{c|}{Mass Challenge} & \multicolumn{2}{c}{Formula Challenge} \\
        \cmidrule(lr){2-3} \cmidrule(lr){4-5}
        \multirow{2}{*}{Train Bin Width (Dalton)} 
        & \multicolumn{2}{c|}{Test Bin Width (Dalton)} 
        & \multicolumn{2}{c}{Test Bin Width (Dalton)} \\
        & 0.1 & 0.01 & 0.1 & 0.01 \\
        \midrule
        0.1  & 0.664 & 0.685 & 0.479 & \textbf{0.490} \\
        0.01 & 0.652 & \textbf{0.697} & 0.471 & 0.485 \\
        \bottomrule
    \end{tabular}
    \label{table:bin_width_ablation}
\end{table}
\section{Additional Experimental Results on NIST'20 Spectra Prediction}
We provide cosine similarity, entropy similarity, and coverage of predicted spectra on NIST'20 clustered by all adduct types, positive adduct types, and [M+H]\textsuperscript{+} adduct type in Table~\ref{tab:spec_sim_hplus}, Table~\ref{tab:spec_sim_pos}, and Table~\ref{tab:spec_sim_all}. \ours outperforms all baseline methods, demonstrating its power on predicting accurate spectra across diverse datasets and instrumental parameters.
\label{section:spec_sim_extra}

\begin{table}[h!]
\caption{Detailed evaluation results on spectral prediction accuracy for [M+H]\textsuperscript{+} adduct on NIST'20 dataset. $\pm$: 95\% CI. Baseline results are adapted from \citet{Wang2025.05.28.656653}.}
\label{tab:spec_sim_hplus}
\centering
\begin{tabular}{lccc}
\toprule
\multirow{2}{*}{Method} & \multicolumn{3}{c}{\textbf{[M+H]\textsuperscript{+}, Random split}} \\
 & Cosine sim. ($\uparrow$) & Entropy sim. ($\uparrow$) & Coverage ($\uparrow$) \\
\midrule
CFM-ID~\citep{wang2021cfm-id4} & 0.470 $\pm$ 0.000 & 0.427 $\pm$ 0.000 & 0.261 $\pm$ 0.000 \\
GrAFF-MS~\citep{murphy2023efficiently} & 0.565 $\pm$ 0.001 & 0.529 $\pm$ 0.001 & 0.740 $\pm$ 0.001 \\
MassFormer~\citep{young2024massformer} & 0.632 $\pm$ 0.001 & 0.605 $\pm$ 0.000 & 0.787 $\pm$ 0.001 \\
FraGNNet~\citep{young2024fragnnet} & 0.717 $\pm$ 0.001 & - & - \\
ICEBERG~1.0~\citep{goldman2024iceberg} & 0.754 $\pm$ 0.002 & 0.700 $\pm$ 0.001 & 0.809 $\pm$ 0.001 \\
ICEBERG~2.0~\citep{Wang2025.05.28.656653} & 0.794 $\pm$ 0.002 & 0.763 $\pm$ 0.001 & 0.856 $\pm$ 0.000 \\
\midrule
\textbf{\ours, w/o CF} & \textbf{0.838 $\pm$ 0.001} & \textbf{0.781 $\pm$ 0.001} & \textbf{0.887 $\pm$ 0.001} \\
\textbf{\ours, w/ CF} & 0.815 $\pm$ 0.003 & 0.752 $\pm$ 0.003 & 0.858 $\pm$ 0.004\\
\midrule
\midrule
\multirow{2}{*}{Method} & \multicolumn{3}{c}{\textbf{[M+H]\textsuperscript{+}, Scaffold split}} \\
& Cosine sim. ($\uparrow$) & Entropy sim. ($\uparrow$) & Coverage ($\uparrow$) \\
\midrule
CFM-ID~\citep{wang2021cfm-id4} & 0.450 $\pm$ 0.000 & 0.401 $\pm$ 0.000 & 0.228 $\pm$ 0.000 \\
GrAFF-MS~\citep{murphy2023efficiently} & 0.470 $\pm$ 0.001 & 0.448 $\pm$ 0.001 & 0.706 $\pm$ 0.001 \\
MassFormer~\citep{young2024massformer} & 0.526 $\pm$ 0.002 & 0.512 $\pm$ 0.002 & 0.742 $\pm$ 0.002 \\
FraGNNet~\citep{young2024fragnnet} & 0.654 $\pm$ 0.003 & - & - \\
ICEBERG~\citep{goldman2024iceberg} & 0.701 $\pm$ 0.001 & 0.641 $\pm$ 0.004 & 0.790 $\pm$ 0.001 \\
ICEBERG~2.0~\citep{Wang2025.05.28.656653} & 0.735 $\pm$ 0.000 & 0.708 $\pm$ 0.000 & 0.847 $\pm$ 0.000 \\
\midrule
\textbf{\ours, w/o CF} & \textbf{0.788 $\pm$ 0.002} & \textbf{0.733 $\pm$ 0.001} & \textbf{0.876 $\pm$ 0.001}\\
\textbf{\ours, w/ CF} & 0.763 $\pm$ 0.020 & 0.697 $\pm$ 0.025 & 0.840 $\pm$ 0.022\\
\bottomrule
\end{tabular}
\footnotetext{As reported in this work, GLACIER-predicted mass spectra are more similar to the experimental spectra than peer methods. As a general trend, entropy similarities tend to score worse compared to cosine similarities, while switching to the other metric does not affect the relative ranking of methods considered. GLACIER-predicted peaks also provide higher coverage (i.e., recall) of experiment peaks. {``Coverage'' denotes the recall in terms of the number of peaks annotated by the model divided by the number of experimental peaks.} Results are reported with 95\% CI on 3 random seeds.}
\end{table}

\begin{table}[h]
\caption{Detailed evaluation results on spectral prediction accuracy for positive mode adducts on NIST'20 dataset. $\pm$: 95\% CI from three random seeds. Baseline results are adapted from \citet{Wang2025.05.28.656653}.}
\label{tab:spec_sim_pos}
\centering
\begin{tabular}{lccc}
\toprule
\multirow{2}{*}{Method} & \multicolumn{3}{c}{\textbf{Positive mode, Random split}} \\
 & Cosine sim. ($\uparrow$) & Entropy sim. ($\uparrow$) & Coverage ($\uparrow$) \\
\midrule
GrAFF-MS~\citep{murphy2023efficiently} & 0.578 $\pm$ 0.001 & 0.533 $\pm$ 0.001 & 0.724 $\pm$ 0.000 \\
MassFormer~\citep{young2024massformer} & 0.640 $\pm$ 0.001 & 0.605 $\pm$ 0.000 & 0.776 $\pm$ 0.001 \\
ICEBERG~1.0~\citep{goldman2024iceberg} & 0.727 $\pm$ 0.002 & 0.664 $\pm$ 0.000 & 0.754 $\pm$ 0.002 \\
ICEBERG~2.0~\citep{Wang2025.05.28.656653} & 0.782 $\pm$ 0.002 & 0.749 $\pm$ 0.001 & 0.837 $\pm$ 0.000 \\
\midrule
\textbf{\ours, w/o CF} & \textbf{0.829 $\pm$ 0.001} & \textbf{0.769 $\pm$ 0.001} & \textbf{0.873 $\pm$ 0.001} \\
\textbf{\ours, w/ CF} & 0.809 $\pm$ 0.002 & 0.743 $\pm$ 0.003 & 0.847 $\pm$ 0.004 \\
\midrule
\midrule
\multirow{2}{*}{Method} & \multicolumn{3}{c}{\textbf{Positive mode, Scaffold split}} \\
 & Cosine sim. ($\uparrow$) & Entropy sim. ($\uparrow$) & Coverage ($\uparrow$) \\
\midrule
GrAFF-MS~\citep{murphy2023efficiently} & 0.477 $\pm$ 0.001 & 0.452 $\pm$ 0.001 & 0.701 $\pm$ 0.001 \\
MassFormer~\citep{young2024massformer} & 0.532 $\pm$ 0.002 & 0.515 $\pm$ 0.002 & 0.740 $\pm$ 0.001 \\
ICEBERG~1.0~\citep{goldman2024iceberg} & 0.698 $\pm$ 0.001 & 0.633 $\pm$ 0.004 & 0.770 $\pm$ 0.001 \\
ICEBERG~2.0~\citep{Wang2025.05.28.656653} & 0.734 $\pm$ 0.000 & 0.707 $\pm$ 0.000 & 0.842 $\pm$ 0.000 \\
\midrule
\textbf{\ours, w/o CF} & \textbf{0.787 $\pm$ 0.002} & \textbf{0.731 $\pm$ 0.001} & \textbf{0.872 $\pm$ 0.001} \\
\textbf{\ours, w/ CF} & 0.758 $\pm$ 0.010 & 0.689 $\pm$ 0.009 & 0.832 $\pm$ 0.008\\
\bottomrule
\end{tabular}
\end{table}

\begin{table}[h]
\caption{Detailed evaluation results on spectral prediction accuracy for both positive and negative mode adducts on NIST'20 dataset. $\pm$: 95\% CI from three random seeds. Baseline results are adapted from \citet{Wang2025.05.28.656653}.}
\label{tab:spec_sim_all}
\centering
{
\begin{tabular}{lccc}
\toprule
\multirow{2}{*}{Method} & \multicolumn{3}{c}{\textbf{Positive+negative mode, Random split}} \\
 & Cosine sim. ($\uparrow$) & Entropy sim. ($\uparrow$) & Coverage ($\uparrow$) \\
\midrule

GrAFF-MS~\citep{murphy2023efficiently} & 0.556 $\pm$ 0.001 & 0.512 $\pm$ 0.001 & 0.710 $\pm$ 0.000 \\
MassFormer~\citep{young2024massformer} & 0.656 $\pm$ 0.001 & 0.610 $\pm$ 0.000 & 0.782 $\pm$ 0.001 \\
ICEBERG~2.0~\citep{Wang2025.05.28.656653} & 0.773 $\pm$ 0.002 & 0.740 $\pm$ 0.002 & 0.834 $\pm$ 0.001\\
\midrule
\textbf{\ours, w/o CF} & \textbf{0.821 $\pm$ 0.001} & \textbf{0.761 $\pm$ 0.001} & \textbf{0.869 $\pm$ 0.000}\\
\textbf{\ours, w/ CF} & 0.802 $\pm$ 0.002 & 0.736 $\pm$ 0.004 & 0.843 $\pm$ 0.003 \\
\midrule
\midrule
\multirow{2}{*}{Method} & \multicolumn{3}{c}{\textbf{Positive+negative mode, Scaffold split}} \\
 & Cosine sim. ($\uparrow$) & Entropy sim. ($\uparrow$) & Coverage ($\uparrow$) \\
\midrule
GrAFF-MS~\citep{murphy2023efficiently} & 0.474 $\pm$ 0.001 & 0.448 $\pm$ 0.001 & 0.692 $\pm$ 0.001 \\
MassFormer~\citep{young2024massformer} & 0.536 $\pm$ 0.001 & 0.516 $\pm$ 0.002 & 0.738 $\pm$ 0.002 \\
ICEBERG~2.0~\citep{Wang2025.05.28.656653} & 0.733 $\pm$ 0.000 & 0.706 $\pm$ 0.000 & 0.837 $\pm$ 0.000 \\
\midrule
\textbf{\ours, w/o CF} & \textbf{0.785 $\pm$ 0.002} & \textbf{0.729 $\pm$ 0.001} & \textbf{0.867 $\pm$ 0.001} \\
\textbf{\ours, w/ CF} & 0.756 $\pm$ 0.010 & 0.687 $\pm$ 0.009 & 0.826 $\pm$ 0.008\\
\bottomrule
\end{tabular}
}
\end{table}
\clearpage
\section{Additional Experimental Results on NIST'20 Retrieval}
\label{section:additional_retrieval}
We provide the retrieval results on the NIST'20 with candidates of the [M+H]\textsuperscript{+} adduct type in Table~\ref{tab:retrieval_acc_hplus} and of positive adduct types in Table~\ref{tab:retrieval_acc_pos}. \ours outperforms all baseline methods, improving top-1 retrieval accuracy on the [M+H]\textsuperscript{+} adduct type from 40.0\% to 53.3\% and 34.0\% to 52.0\% on all positive adduct types. Results of all baseline methods are adapted from \citet{Wang2025.05.28.656653}.

\begin{table}[h]
    \caption{Detailed retrieval accuracy for [M+H]\textsuperscript{+} adduct on NIST'20 dataset. $\pm$: 95\% CI. Baseline results are adapted from \citet{Wang2025.05.28.656653}.
    }
\label{tab:retrieval_acc_hplus}
\resizebox{\textwidth}{!}{
\begin{tabular}{lcccccccccc}
\toprule
& \multicolumn{5}{c}{\textbf{[M+H]\textsuperscript{+}, Random split}} \\
Top-$k$ accuracy & Top-1 & Top-2 & Top-3 & Top-4 & Top-5 \\
\midrule
MetFrag~\citep{ruttkies2016metfrag} & 0.107 $\pm$ 0.000 & 0.193 $\pm$ 0.000 & 0.267 $\pm$ 0.000 & 0.317 $\pm$ 0.000 & 0.375 $\pm$ 0.000 \\
GrAFF-MS~\citep{murphy2023efficiently} & 0.211 $\pm$ 0.004 & 0.365 $\pm$ 0.009 & 0.472 $\pm$ 0.015 & 0.551 $\pm$ 0.013 & 0.608 $\pm$ 0.005 \\
MassFormer~\citep{young2024massformer} & 0.252 $\pm$ 0.001 & 0.422 $\pm$ 0.002 & 0.539 $\pm$ 0.005 & 0.617 $\pm$ 0.007 & 0.675 $\pm$ 0.004 \\
FraGNNet~\citep{young2024fragnnet} & 0.238 & - & 0.504 & - & 0.652 \\
ICEBERG~1.0~\citep{goldman2024iceberg} & 0.251 $\pm$ 0.016 & 0.454 $\pm$ 0.004 & 0.576 $\pm$ 0.006 & 0.654 $\pm$ 0.004 & 0.711 $\pm$ 0.007 \\
ICEBERG~2.0~\citep{Wang2025.05.28.656653} & 0.400 $\pm$ 0.008 & 0.611 $\pm$ 0.013 & 0.721 $\pm$ 0.010 & 0.790 $\pm$ 0.007 & 0.829 $\pm$ 0.006 \\
\midrule
\textbf{\ours (ours, w/ CF)} & \textbf{0.533 $\pm$ 0.009} & \textbf{0.713 $\pm$ 0.001} & \textbf{0.802 $\pm$ 0.001} & \textbf{0.846 $\pm$ 0.002} & \textbf{0.874 $\pm$ 0.005} \\
\midrule
Top-$k$ accuracy & Top-6 & Top-7 & Top-8 & Top-9 & Top-10 \\
\midrule
MetFrag~\citep{ruttkies2016metfrag} & 0.414 $\pm$ 0.000 & 0.457 $\pm$ 0.000 & 0.497 $\pm$ 0.000 & 0.537 $\pm$ 0.000 & 0.567 $\pm$ 0.000 \\
GrAFF-MS~\citep{murphy2023efficiently} & 0.655 $\pm$ 0.006 & 0.690 $\pm$ 0.004 & 0.724 $\pm$ 0.008 & 0.754 $\pm$ 0.010 & 0.776 $\pm$ 0.009 \\
MassFormer~\citep{young2024massformer} & 0.724 $\pm$ 0.005 & 0.761 $\pm$ 0.008 & 0.794 $\pm$ 0.010 & 0.822 $\pm$ 0.005 & 0.843 $\pm$ 0.006 \\
FraGNNet~\citep{young2024fragnnet} & - & - & - & - & 0.831 \\
ICEBERG~1.0~\citep{goldman2024iceberg} & 0.753 $\pm$ 0.007 & 0.783 $\pm$ 0.005 & 0.810 $\pm$ 0.001 & 0.833 $\pm$ 0.007 & 0.850 $\pm$ 0.009 \\
ICEBERG~2.0~\citep{Wang2025.05.28.656653} & 0.859 $\pm$ 0.005 & 0.881 $\pm$ 0.006 & 0.900 $\pm$ 0.009 & 0.913 $\pm$ 0.008 & 0.923 $\pm$ 0.007 \\
\midrule
\textbf{\ours (ours w/ CF)} & \textbf{0.892 $\pm$ 0.006} & \textbf{0.907 $\pm$ 0.005} & \textbf{0.917 $\pm$ 0.004} & \textbf{0.924 $\pm$ 0.005} & \textbf{0.930 $\pm$ 0.004}\\
\midrule
\midrule
& \multicolumn{5}{c}{\textbf{[M+H]\textsuperscript{+}, Scaffold split}} \\
Top-$k$ accuracy & Top-1 & Top-2 & Top-3 & Top-4 & Top-5 \\
\midrule
MetFrag~\citep{ruttkies2016metfrag} & 0.127 $\pm$ 0.000 & 0.225 $\pm$ 0.000 & 0.302 $\pm$ 0.000 & 0.363 $\pm$ 0.000 & 0.420 $\pm$ 0.000 \\
GrAFF-MS~\citep{murphy2023efficiently} & 0.143 $\pm$ 0.004 & 0.270 $\pm$ 0.002 & 0.368 $\pm$ 0.008 & 0.450 $\pm$ 0.006 & 0.518 $\pm$ 0.002 \\
MassFormer~\citep{young2024massformer} & 0.184 $\pm$ 0.008 & 0.328 $\pm$ 0.005 & 0.431 $\pm$ 0.001 & 0.512 $\pm$ 0.005 & 0.575 $\pm$ 0.004 \\
ICEBERG~1.0~\citep{goldman2024iceberg} & 0.229 $\pm$ 0.004 & 0.415 $\pm$ 0.006 & 0.535 $\pm$ 0.008 & 0.617 $\pm$ 0.009 & 0.675 $\pm$ 0.004 \\
ICEBERG~2.0~\citep{Wang2025.05.28.656653} & 0.335 $\pm$ 0.005 & 0.543 $\pm$ 0.004 & 0.669 $\pm$ 0.006 & 0.745 $\pm$ 0.002 & 0.799 $\pm$ 0.004 \\
\midrule
\textbf{\ours (ours, w/ CF)} & \textbf{0.452 $\pm$ 0.005} & \textbf{0.637 $\pm$ 0.005} & \textbf{0.735 $\pm$ 0.010} & \textbf{0.794 $\pm$ 0.011} & \textbf{0.833 $\pm$ 0.010} \\
\midrule
Top-$k$ accuracy & Top-6 & Top-7 & Top-8 & Top-9 & Top-10 \\
\midrule
MetFrag~\citep{ruttkies2016metfrag} & 0.471 $\pm$ 0.000 & 0.510 $\pm$ 0.000 & 0.548 $\pm$ 0.000 & 0.582 $\pm$ 0.000 & 0.612 $\pm$ 0.000 \\
GrAFF-MS~\citep{murphy2023efficiently} & 0.569 $\pm$ 0.002 & 0.614 $\pm$ 0.003 & 0.650 $\pm$ 0.000 & 0.687 $\pm$ 0.002 & 0.716 $\pm$ 0.002 \\
MassFormer~\citep{young2024massformer} & 0.629 $\pm$ 0.004 & 0.674 $\pm$ 0.006 & 0.712 $\pm$ 0.004 & 0.749 $\pm$ 0.004 & 0.775 $\pm$ 0.004 \\
ICEBERG~1.0~\citep{goldman2024iceberg} & 0.721 $\pm$ 0.001 & 0.756 $\pm$ 0.003 & 0.787 $\pm$ 0.001 & 0.812 $\pm$ 0.002 & 0.834 $\pm$ 0.002 \\
ICEBERG~2.0~\citep{Wang2025.05.28.656653} & 0.838 $\pm$ 0.006 & 0.867 $\pm$ 0.005 & \textbf{0.890 $\pm$ 0.006} & \textbf{0.907 $\pm$ 0.009} & \textbf{0.920 $\pm$ 0.009} \\
\midrule
\textbf{\ours (ours, w/ CF)} & \textbf{0.859 $\pm$ 0.011} & \textbf{0.875 $\pm$ 0.009} & 0.889 $\pm$ 0.008 & 0.900 $\pm$ 0.008 & 0.909 $\pm$ 0.010 \\
\bottomrule
\end{tabular}
}
\end{table}

\begin{table}
\caption{Detailed retrieval accuracy for positive mode adducts on NIST'20 dataset. $\pm$: 95\% CI from three random seeds. Baseline results are adapted from \citet{Wang2025.05.28.656653}. 
}
\label{tab:retrieval_acc_pos}
\centering
\resizebox{\textwidth}{!}
{
\begin{tabular}{lccccc}
\toprule
& \multicolumn{5}{c}{\textbf{Positive mode, Random split}} \\
Top-$k$ accuracy & Top-1 & Top-2 & Top-3 & Top-4 & Top-5 \\
\midrule
GrAFF-MS~\citep{murphy2023efficiently} & 0.173 $\pm$ 0.004 & 0.316 $\pm$ 0.003 & 0.417 $\pm$ 0.007 & 0.490 $\pm$ 0.006 & 0.545 $\pm$ 0.000 \\
MassFormer~\citep{young2024massformer} & 0.219 $\pm$ 0.002 & 0.369 $\pm$ 0.004 & 0.476 $\pm$ 0.005 & 0.551 $\pm$ 0.006 & 0.607 $\pm$ 0.005 \\
ICEBERG~1.0~\citep{goldman2024iceberg} & 0.189 $\pm$ 0.012 & 0.375 $\pm$ 0.005 & 0.489 $\pm$ 0.007 & 0.567 $\pm$ 0.005 & 0.623 $\pm$ 0.004 \\
ICEBERG~2.0~\citep{Wang2025.05.28.656653} & 0.340 $\pm$ 0.004 & 0.550 $\pm$ 0.011 & 0.661 $\pm$ 0.005 & 0.735 $\pm$ 0.006 & 0.777 $\pm$ 0.006 \\
\midrule
\textbf{\ours (ours, w/ CF)} & \textbf{0.520 $\pm$ 0.011} & \textbf{0.684 $\pm$ 0.002} & \textbf{0.774 $\pm$ 0.002} & \textbf{0.821 $\pm$ 0.002} & \textbf{0.854 $\pm$ 0.005}\\
\midrule
& \multicolumn{5}{c}{\textbf{Positive mode, Random split}} \\
Top-$k$ accuracy & Top-6 & Top-7 & Top-8 & Top-9 & Top-10 \\
\midrule
GrAFF-MS~\citep{murphy2023efficiently} & 0.591 $\pm$ 0.003 & 0.625 $\pm$ 0.002 & 0.660 $\pm$ 0.004 & 0.690 $\pm$ 0.005 & 0.714 $\pm$ 0.006 \\
MassFormer~\citep{young2024massformer} & 0.654 $\pm$ 0.004 & 0.693 $\pm$ 0.004 & 0.725 $\pm$ 0.004 & 0.755 $\pm$ 0.001 & 0.775 $\pm$ 0.004 \\
ICEBERG~1.0~\citep{goldman2024iceberg} & 0.666 $\pm$ 0.002 & 0.698 $\pm$ 0.002 & 0.725 $\pm$ 0.003 & 0.749 $\pm$ 0.003 & 0.770 $\pm$ 0.002 \\
ICEBERG~2.0~\citep{Wang2025.05.28.656653} & 0.810 $\pm$ 0.006 & 0.834 $\pm$ 0.008 & 0.855 $\pm$ 0.011 & 0.871 $\pm$ 0.008 & 0.885 $\pm$ 0.008 \\
\midrule
\textbf{\ours (ours, w/ CF)} & \textbf{0.876 $\pm$ 0.005} & \textbf{0.893 $\pm$ 0.004} & \textbf{0.905 $\pm$ 0.005} & \textbf{0.915 $\pm$ 0.005} & \textbf{0.922 $\pm$ 0.005}\\
\midrule
\midrule
& \multicolumn{5}{c}{\textbf{Positive mode, Scaffold split}} \\
Top-$k$ accuracy & Top-1 & Top-2 & Top-3 & Top-4 & Top-5 \\
\midrule
GrAFF-MS~\citep{murphy2023efficiently} & 0.138 $\pm$ 0.003 & 0.262 $\pm$ 0.005 & 0.358 $\pm$ 0.008 & 0.437 $\pm$ 0.006 & 0.502 $\pm$ 0.003 \\
MassFormer~\citep{young2024massformer} & 0.179 $\pm$ 0.007 & 0.319 $\pm$ 0.004 & 0.421 $\pm$ 0.003 & 0.499 $\pm$ 0.003 & 0.562 $\pm$ 0.003 \\
ICEBERG~1.0~\citep{goldman2024iceberg} & 0.210 $\pm$ 0.004 & 0.397 $\pm$ 0.006 & 0.518 $\pm$ 0.006 & 0.599 $\pm$ 0.007 & 0.657 $\pm$ 0.002 \\
ICEBERG~2.0~\citep{Wang2025.05.28.656653} & 0.321 $\pm$ 0.006 & 0.525 $\pm$ 0.003 & 0.650 $\pm$ 0.007 & 0.724 $\pm$ 0.005 & 0.778 $\pm$ 0.005 \\
\midrule
\textbf{\ours (ours, w/ CF)} & 
\textbf{0.451 $\pm$ 0.004} & 
\textbf{0.631 $\pm$ 0.004} & 
\textbf{0.729 $\pm$ 0.009} & 
\textbf{0.787 $\pm$ 0.010} & 
\textbf{0.825 $\pm$ 0.009} \\
\midrule
Top-$k$ accuracy & Top-6 & Top-7 & Top-8 & Top-9 & Top-10 \\
\midrule
GrAFF-MS~\citep{murphy2023efficiently} & 0.553 $\pm$ 0.002 & 0.598 $\pm$ 0.003 & 0.633 $\pm$ 0.001 & 0.670 $\pm$ 0.002 & 0.699 $\pm$ 0.001 \\
MassFormer~\citep{young2024massformer} & 0.615 $\pm$ 0.004 & 0.660 $\pm$ 0.007 & 0.698 $\pm$ 0.006 & 0.734 $\pm$ 0.005 & 0.761 $\pm$ 0.006 \\
ICEBERG~1.0~\citep{goldman2024iceberg} & 0.703 $\pm$ 0.001 & 0.738 $\pm$ 0.003 & 0.769 $\pm$ 0.001 & 0.794 $\pm$ 0.003 & 0.817 $\pm$ 0.003 \\
ICEBERG~2.0~\citep{Wang2025.05.28.656653} & 0.818 $\pm$ 0.005 & 0.848 $\pm$ 0.005 & 0.871 $\pm$ 0.004 & 0.889 $\pm$ 0.008 & 0.903 $\pm$ 0.008 \\
\midrule
\textbf{\ours (ours, w/ CF)} & 
\textbf{0.852 $\pm$ 0.010} & 
\textbf{0.869 $\pm$ 0.008} & 
\textbf{0.885 $\pm$ 0.007} & 
\textbf{0.896 $\pm$ 0.008} & 
\textbf{0.907 $\pm$ 0.008} \\
\bottomrule
\end{tabular}
}
\end{table}
\clearpage

\section{Visualization of Predicted Spectra}
\begin{figure}[tbh!]
\includegraphics[width=\linewidth]{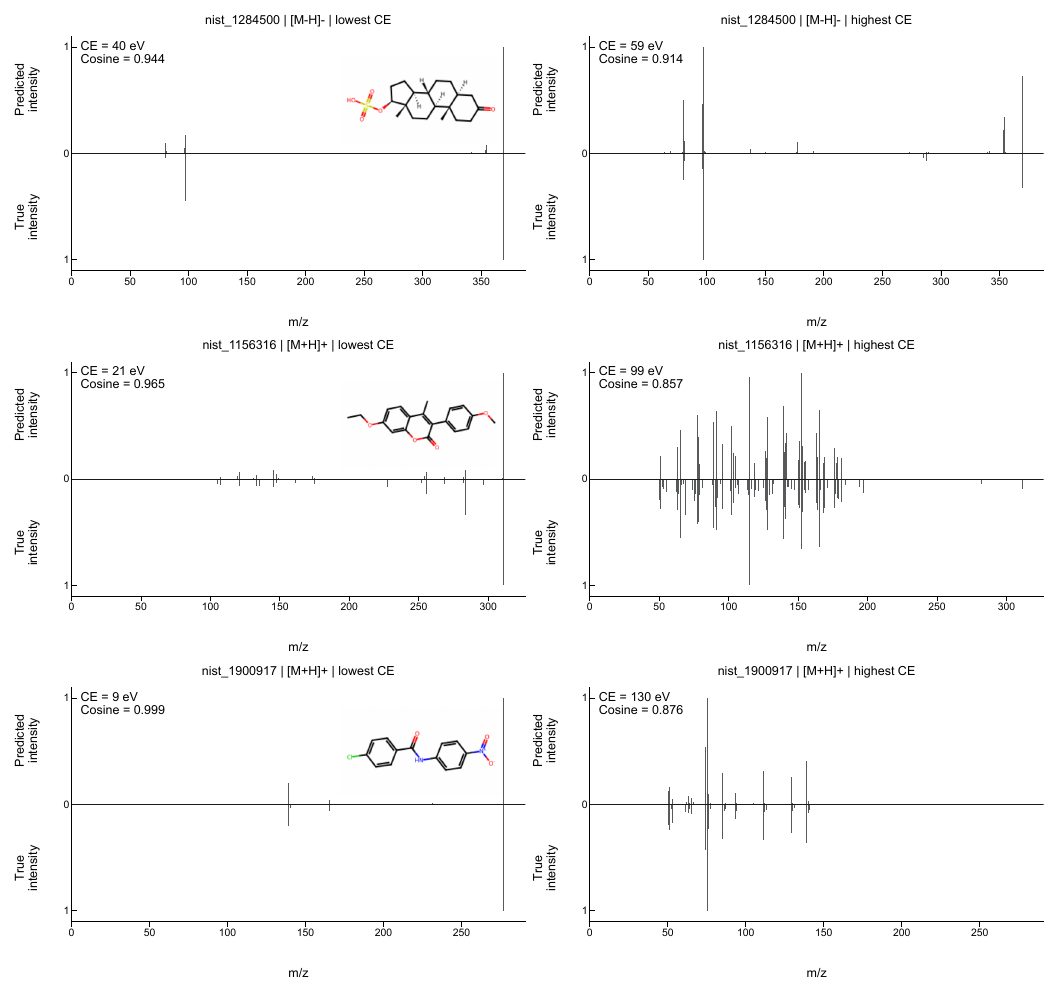}
    \vspace{-20pt}
    \caption{Visualization of predicted spectra from three different molecules with the highest and the lowest collision energy.}    
    \vspace{-5pt}
    \label{fig:spec}
\end{figure}

We sample three different molecules from the NIST'20 dataset and plot the predicted spectra with the highest and lowest collision energy in Figure~\ref{fig:spec} with model trained on the NIST'20 random split with seed set to 1, $d$ set to 3, and $\delta$ set to 0.5.
\section{Visualization of Predicted Break Points} We sample a molecule from the NIST'20 dataset and plot the predicted breakpoints with in Figure~\ref{fig:spec} with model trained on the NIST'20 random split with seed set to 1, $d$ set to 3, and $\delta$ set to 0.5. Predicted breakpoints are labeled in red. 
\begin{figure}[tbh!]
\includegraphics[width=\linewidth]{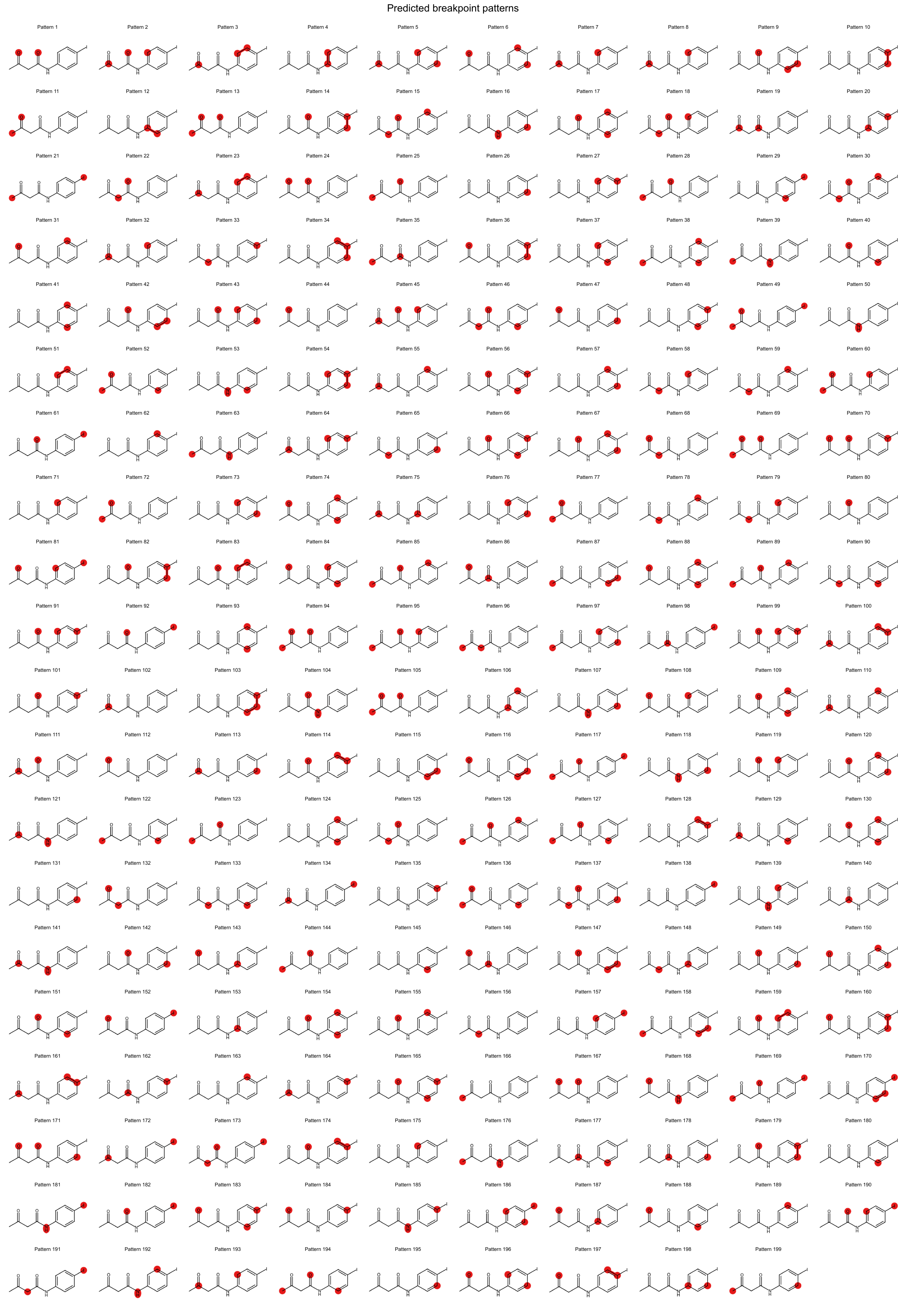}
    \vspace{-20pt}
    \caption{Visualization of predicted breakpoints.}    
    \vspace{-5pt}
    \label{fig:breakpoints}
\end{figure}

\clearpage
\newpage
\input{checklist.tex}

\end{document}

%% file: checklist.tex
\section*{NeurIPS Paper Checklist}

\begin{enumerate}

\item {\bf Claims}
    \item[] Question: Do the main claims made in the abstract and introduction accurately reflect the paper's contributions and scope?
    \item[] Answer: \answerYes{} 
    \item[] Justification: 
    We describe our methods in Section~\ref{section:methods} and we run experiments that verify our claim in  Section~\ref{section:exp}.

\item {\bf Limitations}
    \item[] Question: Does the paper discuss the limitations of the work performed by the authors?
    \item[] Answer: \answerYes{} 
    \item[] Justification: We discuss limitations and future work explicitly in Section~\ref{section:conclusion}.

\item {\bf Theory assumptions and proofs}
    \item[] Question: For each theoretical result, does the paper provide the full set of assumptions and a complete (and correct) proof?
    \item[] Answer: \answerNA{} 
    \item[] Justification: The work in this paper is empirical and does not include theoretical results.

    \item {\bf Experimental result reproducibility}
    \item[] Question: Does the paper fully disclose all the information needed to reproduce the main experimental results of the paper to the extent that it affects the main claims and/or conclusions of the paper (regardless of whether the code and data are provided or not)?
    \item[] Answer: \answerYes{} 
    \item[] Justification: We provide implementation details and hyper-parameters regarding our models, and the datasets we use for evaluation are open-sourced. 

\item {\bf Open access to data and code}
    \item[] Question: Does the paper provide open access to the data and code, with sufficient instructions to faithfully reproduce the main experimental results, as described in supplemental material?
    \item[] Answer: \answerYes{} 
    \item[] Justification: Code is available at \url{https://github.com/coleygroup/ms-pred}.

\item {\bf Experimental setting/details}
    \item[] Question: Does the paper specify all the training and test details (e.g., data splits, hyperparameters, how they were chosen, type of optimizer) necessary to understand the results?
    \item[] Answer: \answerYes{} 
    \item[] Justification: The code is available at \url{https://github.com/coleygroup/ms-pred}.

\item {\bf Experiment statistical significance}
    \item[] Question: Does the paper report error bars suitably and correctly defined or other appropriate information about the statistical significance of the experiments?
    \item[] Answer: \answerYes{} 
    \item[] Justification: All experimental results in Section~\ref{section:exp} include confidence intervals.

\item {\bf Experiments compute resources}
    \item[] Question: For each experiment, does the paper provide sufficient information on the computer resources (type of compute workers, memory, time of execution) needed to reproduce the experiments?
    \item[] Answer: \answerYes{} 
    \item[] Justification: We document the compute resources we use in Section~\ref{section:exp}.
    
\item {\bf Code of ethics}
    \item[] Question: Does the research conducted in the paper conform, in every respect, with the NeurIPS Code of Ethics \url{https://neurips.cc/public/EthicsGuidelines}?
    \item[] Answer: \answerYes{} 
    \item[] Justification: We follow all guidelines described in the protocol.

\item {\bf Broader impacts}
    \item[] Question: Does the paper discuss both potential positive societal impacts and negative societal impacts of the work performed?
    \item[] Answer: \answerYes{} 
    \item[] Justification: We discuss the potential impacts of the paper in Section~\ref{section:conclusion}. We believe this work has a potentially positive societal impact with minimal risk.
    
\item {\bf Safeguards}
    \item[] Question: Does the paper describe safeguards that have been put in place for responsible release of data or models that have a high risk for misuse (e.g., pre-trained language models, image generators, or scraped datasets)?
    \item[] Answer: \answerNA{}{} 
    \item[] Justification: This paper poses no such risk. 

\item {\bf Licenses for existing assets}
    \item[] Question: Are the creators or original owners of assets (e.g., code, data, models), used in the paper, properly credited and are the license and terms of use explicitly mentioned and properly respected?
    \item[] Answer: \answerYes{} 
    \item[] Justification: 
    All assets used in the paper are licensed and properly credited. 

\item {\bf New assets}
    \item[] Question: Are new assets introduced in the paper well documented and is the documentation provided alongside the assets?
    \item[] Answer: \answerYes{} 
    \item[] Justification: All assets introduced in this paper can be accessed in \url{https://github.com/coleygroup/ms-pred}. 

\item {\bf Crowdsourcing and research with human subjects}
    \item[] Question: For crowdsourcing experiments and research with human subjects, does the paper include the full text of instructions given to participants and screenshots, if applicable, as well as details about compensation (if any)? 
    \item[] Answer: \answerNA{} 
    \item[] Justification: 
    This Research does not involve human subjects.

\item {\bf Institutional review board (IRB) approvals or equivalent for research with human subjects}
    \item[] Question: Does the paper describe potential risks incurred by study participants, whether such risks were disclosed to the subjects, and whether Institutional Review Board (IRB) approvals (or an equivalent approval/review based on the requirements of your country or institution) were obtained?
    \item[] Answer: \answerNA{} 
    \item[] Justification: This research does not involve human subjects.

\item {\bf Declaration of LLM usage}
    \item[] Question: Does the paper describe the usage of LLMs if it is an important, original, or non-standard component of the core methods in this research? Note that if the LLM is used only for writing, editing, or formatting purposes and does \emph{not} impact the core methodology, scientific rigor, or originality of the research, declaration is not required.
    \item[] Answer: \answerNA{} 
    \item[] Justification: LLM is only used for paper writing assistance.

\end{enumerate}